\documentclass[conference]{IEEEtran}
\IEEEoverridecommandlockouts
\usepackage{cite}
\usepackage{mathtools}
\usepackage{url}
\usepackage{hyperref}
\usepackage{blindtext}
\usepackage{wrapfig}
\usepackage{enumerate}
\usepackage{footnote}
\usepackage{amsmath,amssymb,amsfonts}
\usepackage{graphicx}
\usepackage{textcomp}
\usepackage{xcolor}
\usepackage{verbatim}
\usepackage{float}
\usepackage{footnote}
\usepackage{mathtools}
\usepackage{multirow}
\usepackage[utf8]{inputenc}
\usepackage{algcompatible}
\usepackage{algorithm}
\usepackage{algorithmicx}
\newcommand\setItemnumber[1]{\setcounter{enumi}{\numexpr#1-1\relax}}
\newcommand{\cond}{\textit{if }}
\newcommand{\for}{\textit{for }}

\def\BibTeX{{\rm B\kern-.05em{\sc i\kern-.025em b}\kern-.08em
    T\kern-.1667em\lower.7ex\hbox{E}\kern-.125emX}}
\makeatletter
\newenvironment{PurelyF}[1][htb]{%
    \renewcommand{\ALG@name}{Improved LTM-MRI algorithm} % Update algorithm name
    
   \begin{algorithm}[#1]%
  }{\end{algorithm}}
\makeatother
\begin{document}
\title{Advancing Robot-Assisted Autism Therapy:\newline A Novel Algorithm for Enhancing Joint Attention Interventions}
\author{
\IEEEauthorblockN{Christian Giannetti}
\textit{Sapienza University of Rome}\\
giannetti.1904342@studenti.uniroma1.it
}
\maketitle
\begin{abstract}
Robot-assisted autism therapy has garnered significant interest within the research community. Recent studies have revealed that using social robots can accelerate the learning process of several skills in areas where autistic children typically show deficits. However, most early research studies conducted interactions via free play. More recent research has demonstrated that robot-mediated autism therapies focusing on core impairments of autism spectrum disorder (e.g., joint attention and imitation) yield better results than unstructured interactions. This paper aims to systematically review the most relevant findings concerning the application of social robotics to joint attention tasks, a cardinal feature of autism spectrum disorder that significantly influences the neurodevelopmental trajectory of autistic children. Initially, we define autism spectrum disorder and explore its societal implications. Following this, we examine the need for technological aid and the potentialities of robot-assisted autism therapy. We then define joint attention and highlight its crucial role in children's social and cognitive development. Subsequently, we analyze the importance of structured interactions and the role of selecting the optimal robot for specific tasks. This is followed by a comparative analysis of the works reviewed earlier, presenting an in-depth examination of two distinct formal models employed to design the prompts and reward system that enables the robot to adapt to children's responses. These models are critically compared to highlight their strengths and limitations. Next, we introduce a novel algorithm to address the identified limitations, integrating interactive environmental factors and a more sophisticated prompting and reward system. Finally, we propose further research directions, discuss the most relevant open questions, and draw conclusions regarding the effectiveness of social robotics in the medical treatment of autism spectrum disorders.
\end{abstract}
%%%%%%%%%%%%%%%%%%%%%%%%%%%%%%%%%%%%%%%%%%%%%%%%%%%%%%%%%%%%%%%%%%%%%%%%%%%%
\section{Introduction: Exploring the Multifaceted Nature of Autism Spectrum Disorders}
\subsection{The Societal and Economic Impact of Autism Spectrum Disorders}
\label{socialproblem}
Autism can be defined as a pervasive neurodevelopmental disorder that affects how individuals perceive the world and interact with others. The core symptoms of autism spectrum disorder (ASD) include social communication difficulties, limited social interactions, and repetitive and stereotyped behaviors. According to the Centers for Disease Control and Prevention (CDC), the prevalence of autism has risen to one in every sixty-eight children \cite{amaral2017examining}. Several studies indicate that the number of children diagnosed with autism continues to grow over the years. Considering that the estimated cost of supporting an individual with ASD over their lifetime amounts to 2.4 million dollars, autism is undoubtedly a significant social issue. Hence, developing innovative approaches to improve the effectiveness of medical assistance in this field is of primary importance. Moreover, as observed by Warren et al. \cite{warren2011therapies} the most effective treatment for improving the medical condition of autistic children is intensive behavioral intervention. Such therapies typically require several hours from qualified therapists and medical personnel, often unavailable in many communities or beyond the financial resources of families and service systems. Consequently, high-quality behavioral intervention is frequently inaccessible to the vast autistic population (\cite{knapp2009economic}, \cite{chasson2007cost}). Considering these factors, the U.S. Department of Health and Human Services has identified the practical identification and treatment of autism spectrum disorders as a public health emergency \cite{interagency2013interagency}.

\subsection{A Different Way to Communicate and Perceive the World}
Concerning communication, it has been observed that children with ASD can spontaneously communicate. However, the learning process demands much more adult involvement and often the use of particular techniques to build dialogue, as highlighted by Coeckelbergh et al. \cite{coeckelbergh2016survey} and Field et al. \cite{field2001children}. Regarding emotion recognition and understanding based on facial expressions, these studies show that people with ASD develop different emotion recognition strategies compared to their neurotypical peers. However, they mimic emotional expressions as well as typically developing individuals. Individuals with autism often struggle to understand social signals, which affects their sense of security. This lack of security generally leads to a desperate search for stability, often manifested in a tendency to organize the world to avoid unknown occurrences.
Consequently, autistic people frequently adopt several compulsive practices and routines to predict every eventuality in their daily lives. Regarding sensory perception, the research community increasingly recognizes that autistic individuals' senses function differently. For example, during medical examinations, it is often observed that a child with ASD has normal hearing but responds differently to stimuli (e.g., not responding to their name or covering their ears in response to loud noises). Another typical characteristic of autistic people is exceptionally high sensitivity to physical contact. Children with ASD require various stimulations to develop an acceptable tolerance to physical contact. They usually need adequate resting time to process perceived stimuli, as an overload of stimuli drives autistic children to avoid social interactions with inherently unpredictable humans.
\subsection{The Need for Innovative Multidisciplinary Approaches}
Autism Spectrum Disorder can often be comorbid\footnote{We define comorbidity as the simultaneous presence of two or more diseases or medical conditions in a patient \cite{stedman1920stedman}} with some level of Intellectual Disability (ID) \cite{edition2013diagnostic}. As Baio et al. pointed out \cite{autism2012prevalence}, 54\% of children with ASD have an Intellectual Quotient (IQ) below 85. This clinical condition is characterized by severe deficiencies in both intellectual capacities and adaptive behavior, resulting in difficulties in reasoning, learning, problem-solving, communication, and social skills. Multiple research teams have reported imitation deficits in children with ASD (\cite{edwards2014meta, williams2004systematic}). Due to intellectual disabilities, therapeutic interventions become more complex, highlighting the need for technological aid \cite{diehl2012clinical}.
\subsection{The Need for Technological Aid}
Based on the statistical report mentioned in section \ref{socialproblem}, the number of children diagnosed with autism has unquestionably increased over time. However, the number of medical personnel and occupational therapists remains inadequate. The shortage of occupational therapists poses a significant challenge to implementing effective intervention programs for individuals with ASD. On average, training qualified medical doctors takes seven years, while training certified occupational therapists and child-care personnel take four and three years, respectively. Consequently, developing robotic devices for autism intervention programs is crucial to overcoming this shortage and increasing the capacity of medically trained personnel to provide intervention programs for autistic children. It is observed that a medically-trained person can, on average, provide help to a single autistic child at a time. Using social robots in therapies and interventions increases the number of individuals treated simultaneously, resulting in more effective and cost-optimized therapies. To date, the design of social robots for providing Robot-Assisted Autism Therapies (RAAT) has gained growing attention in research communities worldwide. However, despite the increasing availability of robotics tools developed for interventions with autistic children, there are no standardized references or benchmarks concerning their use for addressing significant impairments in autism. Furthermore, due to the novelty of these therapies, robotics research in this field lacks reports on the effectiveness of long-term intervention programs. Hence, the objective of this review is twofold: first, to acknowledge the gaps existing in the field of robotics research concerning the applications considered in this paper; second, to provide a concise but critical analysis for the reader.
\subsection{Proposed Approach to the Analysis}
\label{txt:proposedApproach}
In this paper, we will pursue the following approach. First, we will individually review each paper included in the analysis to provide a comprehensive overview of the employed methodologies, achieved results, and intrinsic limitations. Each paper's review will focus on its scientific contribution to the state of the art and the limitations related to the design choices made. While differing in implementation and design choices, some works share underlying logic and can be read together. In such cases, we will provide an additional critical analysis that focuses on how a specific study overcomes the limitations of another and to what extent. The ultimate goal of this critical reading is to provide a robust review that may promote intersections between multiple research studies and encourage further research in the field.
\section{Why Use Social Robots? The Potential of Robot-Assisted Autism Therapy}
With the modern advancements in social robotics, human-like robots have begun to be employed in several applications, including the healthcare field. In this respect, several studies notify there is increasing anecdotal evidence that using social robotics to treat individuals affected by autism spectrum disorders may lead to satisfying results (e.g., \cite{pennisi2016autism}, \cite{diehl2012clinical_review}, \cite{diehl2014clinical}, \cite{ricks2010trends}, \cite{scassellati2012robots}). The following items summarize the primary reasons why robot-assisted autism therapy is particularly effective.
\begin{itemize}
    \item \textit{Avoiding sensory and emotional overstimulation} \newline Evidence-based researches \cite{schreibman2015naturalistic} have proved that developmental intervention was effective for autistic individuals. However, people with ASD may often find it challenging to maintain high motivation and concentration for human intervention \cite{warren2015can}. This is because human beings' dynamic facial features and expressions may affect the intensive sensory processing of autistic individuals. In other terms, the dynamic nature of human beings' facial features and expressions are likely to induce sensory and emotional overstimulation \cite{johnson2007identification}. Consequently, such a phenomenon hinders a proper learning process as autistic individuals tend to evade sensory stimulations actively and instead focus on predictable basic features. Several robots have been developed to address this issue. An eloquent example consists of the Dream Robot, a zoomorphic robot used for robot-assisted autism therapy \cite{szymona2021robot}. Specifically, compared to the therapist's face, the limited mimicry of the robot's muzzle has proved to be more understandable and predictable. This latter characteristic allowed the robot's expressions to be in the zone of proximal development of autistic children, allowing them to learn how to understand several emotions. In this respect, the robot becomes a bridge to a better comprehension of neurotypical people's emotions.
    \item \textit{Designing structured and standardized intervention} \newline Robots enable medical personnel to design a more structured and standardized intervention thanks to the possibility of controlling and replicating a scene with continuous and genuine conversations. In fact, unlike human beings, robots, which by definition operate within predictable and lawful systems, can build a profoundly structured learning environment for autistic individuals. A structured learning environment enables therapists to train autistic people to focus on relevant stimuli and discern the most significant social signals among the many exchanged during social interactions. Furthermore, a no less important benefit is the possibility of minimizing the amount of stress experienced by the patient, which profoundly influences his motivation to actively participate in medical treatments. As pointed out by Scassellati et al. \cite{scassellati2007social}, and later by Feil et al. \cite{feil2009toward}, the advantage of employing structured interactions with humanoid robots consists of the opportunity to create structured social circumstances in which specific social behaviors may occur, resulting in more repeatable, measurable and effective therapies.
    \label{txt:structuredInt_general}
    \item \textit{Higher degree of task engagement} \newline Various research studies  (\cite{diehl2012clinical_review},\cite{feil2009toward},\cite{schadenberg2020differences},\cite{costescu2015reversal}) have revealed that individuals affected by autism spectrum disorders often reach a greater degree of task engagement by interacting with robots than through interactions with humans. Precisely, they are more actively involved in the interaction when facing a robot than their human counterparts (\cite{kumazaki2018can},\cite{kumazaki2018impact},\cite{kumazaki2019brief}). Generally, humans feel a higher affinity degree towards other persons than that to artificial objects. On the other hand, autistic individuals have neither bias toward humans nor repulsion toward artificial objects (in contrast to actual humans). In fact, Nakano et al. \cite{nakano2010atypical} has pointed out that, in most cases, autistic people manifest behaviors toward robots that neurotypical people have toward humans. It follows that employing a robot for medical treatment that requires the patient to actively participate in therapies becomes particularly effective since it exploits the inner affinity that autistic people exhibit towards robots.
    \item \textit{Treatments' long-term effectiveness} \newline
    Therapies for autistic individuals require patience from the therapists and long-term interventions. The design and implementation of long-term therapy is subject to multiple variables which may impede its progress (e.g., long-term unavailability of the medical personnel or occupational therapists, et cetera). On the contrary, a robot, by definition, can perform therapeutic interventions consistently, which is indeed a significant advantage. Concerning the long-term effects of robot-assisted autism therapies, the results have been proved to be promising. For instance, Kumazaki et al. \cite{kumazaki2019role} reported favorable carryover effects after job-interview training using an android robot. Specifically, a 1-year follow-up after the therapy has revealed that at least half of the participants in the research had been hired after passing job-interview examinations. Considering the employment difficulties for people affected by ASD, this is unquestionably an exceptionally positive result.
\end{itemize}
\section{Joint Attention: An Overview}
Joint attention is defined as an early-developing social-communicative skill through which an individual coordinates attention with a social partner or environment's feature by the acts of eye-gazing, pointing, or other verbal or non-verbal signs. Research has revealed that an accurate breakdown of this skill consists of the response to joint attention (RJA) and initiation of joint attention (IJA). The former identifies the ability to shift visual attention by following the cues like pointing and gaze, while the latter refers to seeking others' attention using one's own gestures and gaze \cite{billeci2017integrated}. Joint attention is a fundamental basis for developing communicative abilities and early social and cognitive skills (\cite{delinicolas2007joint}, \cite{schertz2013effects}). It has been observed that early interventions that promote joint attention play a crucial role in medical treatments because these strategies stimulate children to learn from their environment and change their developmental trajectories (\cite{kasari2010randomized},\cite{poon2012extent}). The application of social robots to improve joint attention children's capabilities is gaining increasing attention in the research community. To be involved in a joint attention task correctly, children must orient themselves toward their social partners and quickly shift attention between social and non-social stimuli during the interaction's evolution. As observed by Anzalone et al. \cite{anzalone2014children}, autistic children require an adequate interaction partner to practice and develop joint attention capabilities. However, children with autism spectrum disorders usually do not manifest sustained motivation to interact with an interaction partner, making it difficult for the medical personnel to sustain extensive therapeutic interactions. Since robots are immune to any kind of fatigue, several research teams have taken an interest in employing social robots to tackle this crucial task.
\section{Can Human-Robot Interaction Improve Joint Attention Skills?}
\subsection{Introduction} 
Interventions for ASD individuals widely vary in terms of scope and methodology. Generally, these approaches aim to couple the benefits of developmental and discrete trial procedures through intensive graduated methods of prompts used in game-like and interactional frameworks. As Yoder and McDuffie have observed \cite{yoder2006treatment}, the most promising approaches for improving autistic individuals' core skills are characterized by an interactive nature to stimulate the ASD children's active participation. In fact, the early intervention literature to date \cite{poon2012extent} has revealed that social communication intervention approaches are more effective when children exhibit sustained engagement with various objects, which can be utilized within intrinsically motivating contexts, and when an adequate adjustment in case of modest improvements and shifts can be integrated and employed over time. In other terms, to maximize therapies' effectiveness is crucial to continuously adapt the therapies' contents and the environment's settings to every patient's needs. 
\hfill \break \indent
The first work, object of the analysis, has been conducted by Warren et al. and reported in the companion paper \cite{warren2015can}. The research team has investigated the possibility of integrating a social robot into an interactive intervention environment to improve early joint attention skills in children affected by autism spectrum disorders. Specifically, the project's goal was to explore throughout several sessions the potentialities of an innovative adaptive robot-mediated architecture able to administer joint attention prompts through a humanoid robot and eventually activate features of the intervention environment to improve the therapy's effectiveness. The main merit of this study has been to steer the way to the application of modern robotics technologies to joint attention therapies for autistic children. Precisely, the project aimed to devise and empirically assess the feasibility, usability, and preliminary efficacy of an adaptive interactive robot-mediated system capable of improving joint attention capabilities' performance in children affected by autism spectrum disorders.
\subsection{System Structure}
\begin{figure}[htbp]
    \centering
    {\includegraphics[width=1\linewidth]{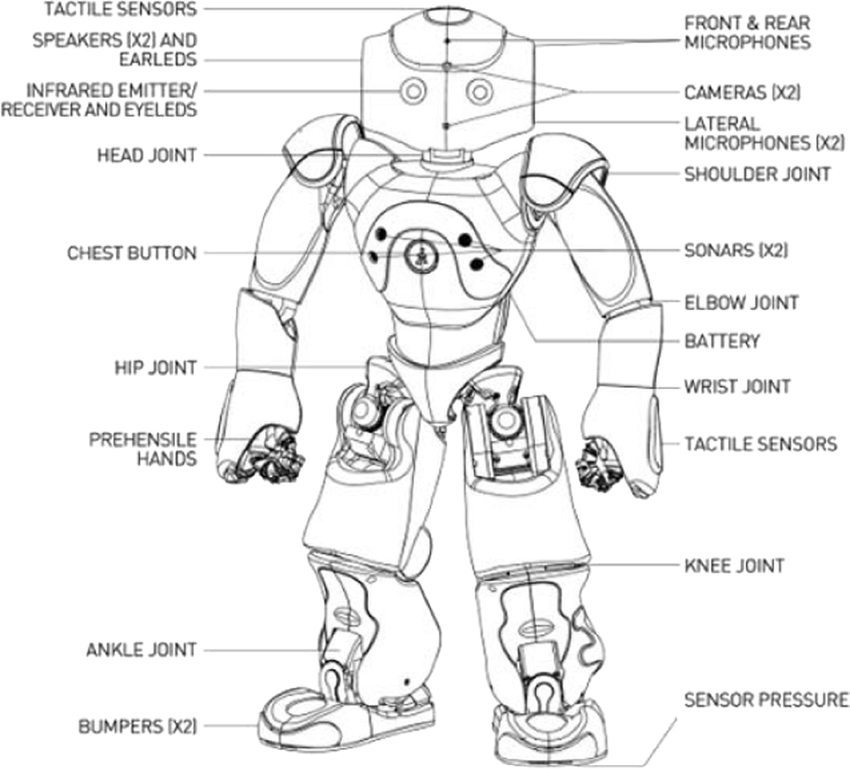}}
    \caption{Breakdown of the NAO robot.}
    \label{fig:im1}
\end{figure}

The system employed in this research study was implemented as a component-based distributed architecture capable of exploiting the network to achieve real-time interaction. Precisely, the system components included:
\begin{enumerate}
    \item A humanoid robot to provide joint attention prompts. Precisely, the employed robot was NAO, depicted in figure \ref{fig:im1}. This robot is a commercially available child-sized plastic-bodied humanoid robot (58 cm tall, 4.3 kg), whose employment is common in several recent applications for autistic children \cite{bekele2014pilot}.
    \item Two target monitors that could eventually be activated if children looked toward them in response to a joint attention prompt.
    \item An eye tracker and linked camera apparatus to observe time spent looking at the robot and evaluate the task performance's correctness.
    \item A Wizard-of-Oz style human-control system, used to identify correct performance. Specifically, the term Wizard-of-Oz is generally employed within the field of human-computer interaction to refer to systems that seem to run autonomously to the participant but are partially operated by unseen human administrators.
\end{enumerate}
%%%%%%%%%%%%%%%%%%%%%%%%%%%%%%%%%
%%%%%%%%%%%%%%%%%%%%%%%%%%%%%%%%%
\subsection{Trial Characteristics}
Each trial comprised up to six possible prompt levels. For each trial, the system randomly selects as target the left or right monitor for the entire trial's duration. In the research study currently presented the robot could only turn its head or turn while pointing to the selected target. After executing each prompt, a 7 seconds response time window was set. Precisely, the interaction was considered a "target hit" if the participant responded correctly (i.e., turned to look) to the selected target within this 7 seconds window. After each prompt's execution, the robot returned to a neutral position (i.e., facing the participant). A technician constantly monitored the participant's performance through direct observation and the calibrated eye-tracking system. 
\subsection{Prompts and Reward Provision System}
\label{txt:LTM}
\begin{figure}[htbp]
    \centering
    {\includegraphics[width=1\linewidth]{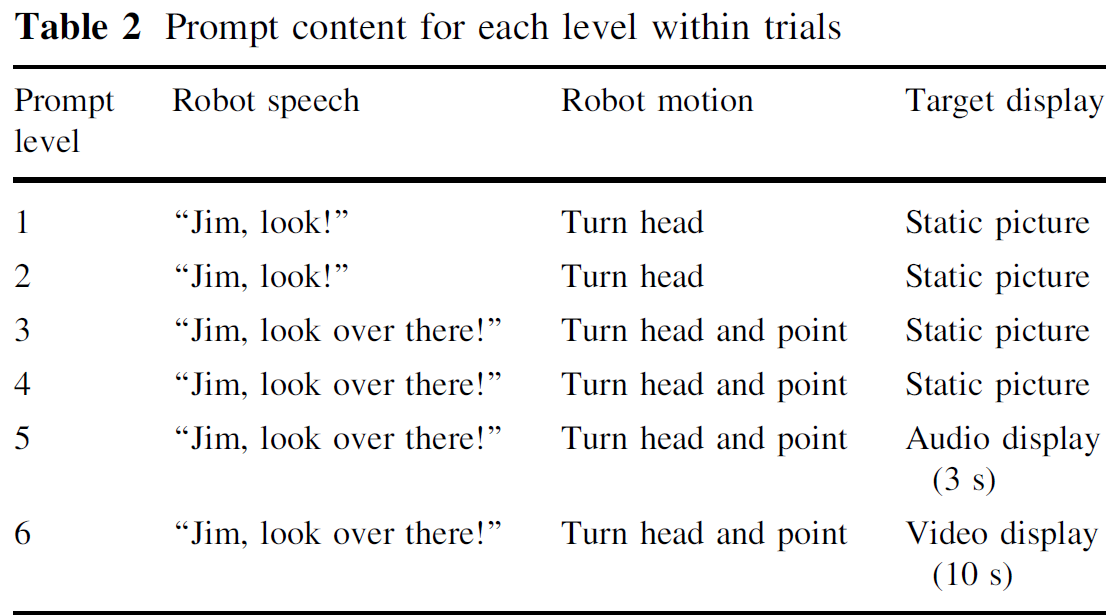}}
    \label{fig:im2}
\end{figure}

The underlying logic that regulates the interaction consists of a hierarchy of prompts given by the robot. The research team has employed a least-to-most prompt (LTM) hierarchy, a standard convention in autism spectrum disorders intervention, which basically provides support to the learner only in case of necessity. The method enables the participant to respond independently at the outset of the task, ensuring opportunities for successful execution and reinforcement at baseline, and only provides increasing support when the child has been given an opportunity to exhibit independent abilities. More precisely, if a target hit was registered, the system triggered a reward (a clip from a children's cartoon was displayed on the target monitor) and started the subsequent trial. If the participant did not follow the robot's instruction within the time window, the system registered the unsuccessful response and progressed to the next level of prompting until all six prompts were administered. Figure \ref{fig:im2} explicitly defines the prompt hierarchy employed in the analyzed paper. 
\label{txt:structuredInt_paper1}
%%%%%%%%%%%%%%%%%%%%%%%%%%%%%%%%%
%%%%%%%%%%%%%%%%%%%%%%%%%%%%%%%%%
\subsection{Results}
The primary purpose of the presented study was to empirically test children's performance in response to within-system joint attention prompts over a series of sessions. The secondary objective was to estimate the attention directed towards the robot over time. To assess these results, the research team has investigated whether children would be responding to the humanoid robot at lower levels of prompting within the hierarchy from baseline to outcome, and also if children's attention towards the robot would diminish over time. Hence, two major indicators have been monitored: the target hit rates to estimate the differences from baseline to final performances and the time spent looking toward the robot across similar time frames. The following paragraphs summarize the reported results.
\subsubsection{Target hit rate}
\begin{figure}[htbp]
    \centering
    {\includegraphics[width=1\linewidth]{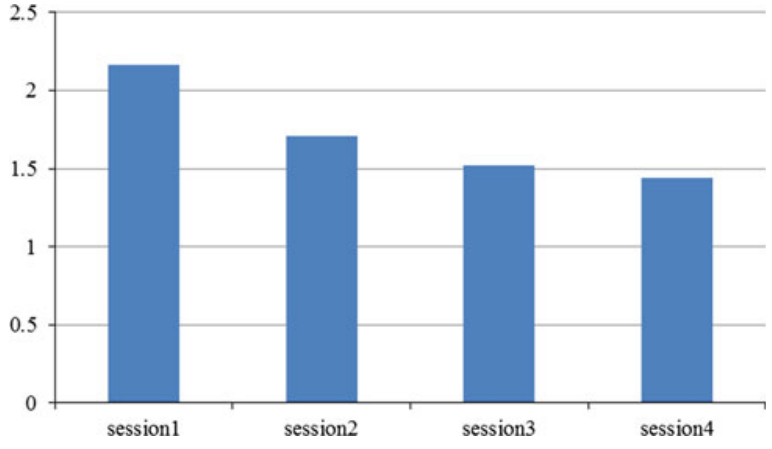}}
    \caption{Average prompt level of target hits across sessions.}
    \label{fig:im3}
\end{figure}

\begin{figure}[htbp]
    \centering
    {\includegraphics[width=1\linewidth]{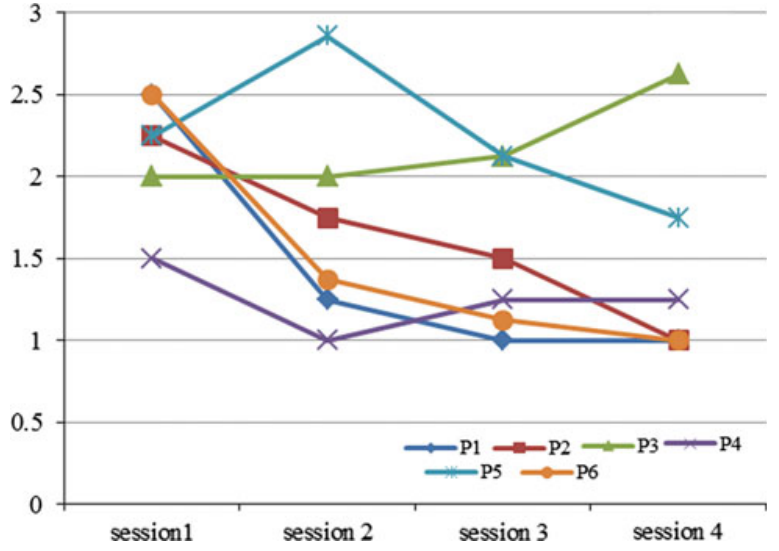}}
    \caption{Average participant prompt level for target hit.}
    \label{fig:im4}
\end{figure}

The results concerning the target hit rate across all sessions and participants were particularly satisfying: 99.48\% of the 32 trials ended in a target hit. Another indicator worth monitoring is the average prompt level before participants hit the target, representing the number of prompts needed before achieving a successful target hit. Figure \ref{fig:im3} graphically represents how the average prompt level of target decreased across sessions. In fact, in session 1, the average target hit prompt level was 2.17 with a standard deviation of SD = 1.49, while, in session 4, the average target hit prompt level has fallen to 1.44 (SD = 1.05). 
%%%%%%%%%%%%%%%%%%%%%%%%%%%%%%%%%
%%%%%%%%%%%%%%%%%%%%%%%%%%%%%%%%%
\hfill \break \indent
Furthermore, a two-sided Wilcoxon rank-sum test has registered that the median difference between session 1 and session 4 was statistically significant ($\rho = 0.003$). Figure \ref{fig:im4} shows children's individual performance over time, indicating that five of the six participating children manifested lower average levels of prompt level target hit across sessions.
%%%%%%%%%%%%%%%%%%%%%%%%%%%%%%%%%
%%%%%%%%%%%%%%%%%%%%%%%%%%%%%%%%%
\hfill \break \indent
\subsubsection{Eye gaze patterns}
Concerning the eye gaze patterns, the research group has carried out two kinds of analysis. The eye gaze patterns have been monitored, firstly, across the whole session, i.e., from the start of the first prompt to the end of the session), and, secondly, within the 7 s response time window across all prompts within a trial. Furthermore, the average time required by the participants to look at the robot across all sessions was 14.75\% of the total experiment time. In addition, within the 7 s window, the average time the participants looked at the robot across all sessions was 24.80\% of the total experiment time. Considering how challenging the task is, these results must indeed be considered satisfying. 
\subsubsection{Discussion of the Results}
The results show that the autistic children from the sample documented improved performance across sessions and manifested sustained interest towards the facilitator robot throughout interactions. These two findings are particularly promising with the ultimate objective of employing this technology in clinical settings. Specifically, children in the sample could often respond correctly to prompts delivered by the humanoid robot within the standardized protocol borrowed from Bekele et al. \cite{bekele2014pilot}. Moreover, the participants also spent a meaningful part of the trial sessions looking at the humanoid robot. These latter results suggest that young children with autism spectrum disorders manifest attentional preferences toward robotic interactions over brief time intervals. In addition, the research group has also demonstrated that children could document increased performance over time in an essential core social communication ability and area of deficit (i.e., response to joint attention) and that, throughout sessions, children maintained an interest in the humanoid robot platform.
\hfill \break \indent
Collectively, these results indicate that robot-mediated systems provided with mechanisms for effectively promoting participants toward correct orientation to various targets could take advantage of non-social attention preferences for autistic individuals to significantly improve capabilities related to coordinated attention over time. However, to achieve these results is undoubtedly necessary to develop a behaviorally sophisticated prompting and reinforcement system, absent in the presented research study.
\subsubsection{Limitations}
This subsection aims to highlight the numerous and significant methodological limitations presented by this research.
\begin{itemize}
    \item \textit{Small population sample and limited time frame of interaction:}
    One of the most impactful limits of this study is undoubtedly the small sample size considered and the limited time frame of interaction. Moreover, the authors have reported a substantial lag between participant assessment and enrollment, resulting in a not fully understanding the sample population in this study.
    \item \textit{Technological limitations:} In most cases, humanoid robots are not as capable of performing sophisticated actions, decoding responses from individuals, and adjusting their behavior within social environments as their human counterparts. Specifically, though NAO is a state-of-the-art commercial humanoid robot, its interaction capabilities exhibit profound limitations. For instance, it creates noise while moving its limbs, limitations in terms of flexibility and degrees of freedom results in less precise gestural motions, and its embedded vocalizations have non-negligible limits due to its basic text-to-speech capabilities. 
    \label{txt:techlimit1}
    \newline
    Another critical technical limitation of the presented study is the employment of a human inside the robotic system loop. Although the presence of a qualified therapist within the loop provides unquestionable benefits in terms of tolerability (i.e., it is more likely that all children complete the protocol), the Wizard-of-Oz paradigms bring further human resource burdens. In this respect, the development of non-invasive technologies for gaze capturing and monitoring would be decisive for realistically deploying the presented system without facing significant fail rates associated with a worn head tracking device\footnote{This phenomenon is mainly caused by the extreme sensibility and discomfort that being touched generates in autistic individuals.} or requiring costly technological solutions (e.g., multiple integrated systems of eye-trackers). Moreover, these systems would promote understanding and tracking visual attention patterns with potential application toward more robust systems.
    \item \textit{Generalization of the improved skill to human-human interaction:}
    Although several research studies have proved that robotic stimuli and systems positively impact and preferentially capture and shift attention, it has not adequately investigated how their performance would compare to instructions provided by their human counterparts.  While the research's result seems promising, the employed methodology profoundly limits the possibility to realistically estimate the value and ultimate clinical utility of the proposed system as consistently applied to autistic children. In fact, robot-mediated systems' potential success and clinical utility depend on their capabilities to stimulate and foster significant changes in core skills linked to dynamic neurodevelopmentally appropriate learning across environments. Although the research team has obtained positive results in the short period, the improvements have not been systematically compared to the learning results achieved through other methods nor investigated if the learned joint attention improvement can be effectively generalized to human-human interactions. In this respect, questions concerning whether the proposed system could be a viable intervention paradigm remain open.
\end{itemize}
\section{Comparative Reading: Choosing the Optimal Robot and the Need for Structured Interactions}
\subsection{Introduction}
The second research study reviewed has been conducted by Kumazaki et al. \cite{kumazaki2018impact}. The primary purpose of the research was to examine whether the robot was more effective in promoting joint attention than a human agent in interventions designed for autistic children. Secondly, the research team has verified whether the skill improvements generalized to human-human robot interactions, i.e., if children with autism spectrum disorder exhibited improvements in joint attention tasks during human interactions after interacting with the robot. Thirdly, the third goal of the research study was to test whether the robot was more effective in facilitating joint attention in autistic children than in neurotypical children. Precisely, the research group has advanced the following hypothesis:
\begin{enumerate}[i)]
    \item Autistic children would have performed better in joint attention tasks under the robot condition than under the human agent condition.
    \item Autistic children would have exhibited improvement in joint attention tasks with a human after interacting with CommU, i.e., the acquired skills after interacting with the robot generalize to human-human interactions.
    \item The facilitative effect of joint attention owing to robot intervention would be more effective in autistic children than in their neurotypical peers, i.e., the robot therapy takes advantage of autistic children's non-social preferences.
\end{enumerate}
\subsection{System Structure}
The system employed in this research study was built as a component-based distributed structure that exploits methods conventionally employed in robotics studies to achieve real-time interaction. Specifically, the system components included:
\begin{enumerate}
    \item A humanoid robot that provided joint attention prompts. Precisely, the robot selected as a joint attention facilitator was the communication robot “CommU.” It was revealed to be particularly suitable for this task because it has clear eyes and can turn them. Since eye contact is a fundamental social ability that autistic children often lack, CommU’s clear eyes enable children to easily identify and understand the communication signals, finally facilitating joint attention.
    \item Two images were disposed respectively on the left and right sides of the participant. They were employed as the foci of attention by the system and substituted at each session to avoid participants' loss of interest.
    \item A Wizard-of-Oz style human-control system, employed to evaluate the performance's correctness. As is often the case for systems conventionally applied in studies related to autism spectrum disorders, the robot was commanded by the researchers, who were not visible during the trial, to elicit the impression that the robot behaved and answered autonomously.
\end{enumerate}
%%%%%%%%%%%%%%%%%%%%%%%%%%%%%%%%%
\begin{figure}[htbp]
    \centering
    {\includegraphics[width=1\linewidth]{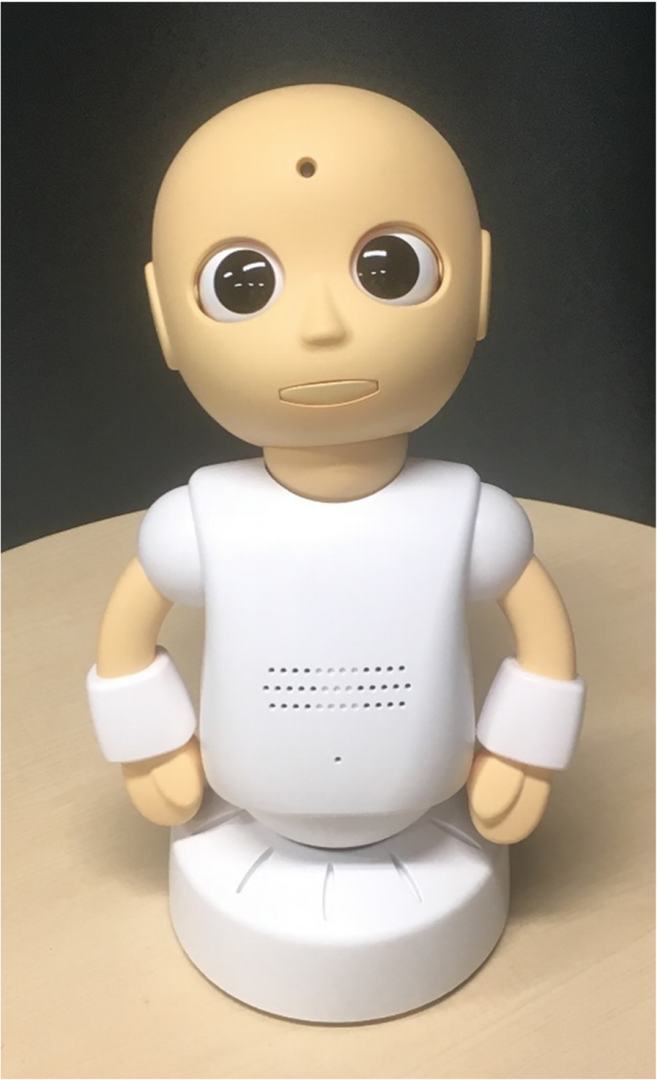}}
    \caption{CommU social robot.}
    \label{fig:im5}
\end{figure}
%%%%%%%%%%%%%%%%%%%%%%%%%%%%%%%%%
\subsection{Trial Characteristics}
The autistic children and the typically-developed children were randomly allocated either to the robot or the control group. Both the participants completed a sequence of three consecutive interaction conditions. In in the robotic intervention group, the participants interacted with \textit{human A, CommU,} and \textit{human A.} On the contrary, in the control group, the participants interacted with \textit{human A, human B,} and \textit{human A.} The entire interaction time for all the participants lasted approximately 15 minutes. Both the human agent and CommU followed a specific interview script and protocol during each interaction to ensure a structured interaction procedure. Across the sessions, the interaction contents were slightly modified to promote engagement but reflected the same basic structure. Depending on the group to which the participants belonged during the trial, the human agent or CommU tried to induce joint attention by alternatively looking toward the child for 1 second and then in the image's direction on the participant's left side for 3 seconds. Consequently, the human agent or CommU gazed again toward the child for 1 second and then the other image's direction on the participant's right side for 3 seconds using the identical scheme. The achievement of joint attention, i.e., a target hit, was determined as a participant responding (i.e., turning to gaze at) to the correct target within a 3 seconds window. Regardless of the participant's response, the human agent or robot returned to a neutral position, i.e., standing straight and facing the participant after every prompt. Each target hit was counted only if the child turned his head and/or eyes towards the target's direction without completely turning his body, following the social prompt. Specifically, in this case, the joint attention event was rated 1 (success), otherwise 0 (failure).
\subsection{Results}
\begin{figure}[htbp]
    \centering
    {\includegraphics[width=1\linewidth]{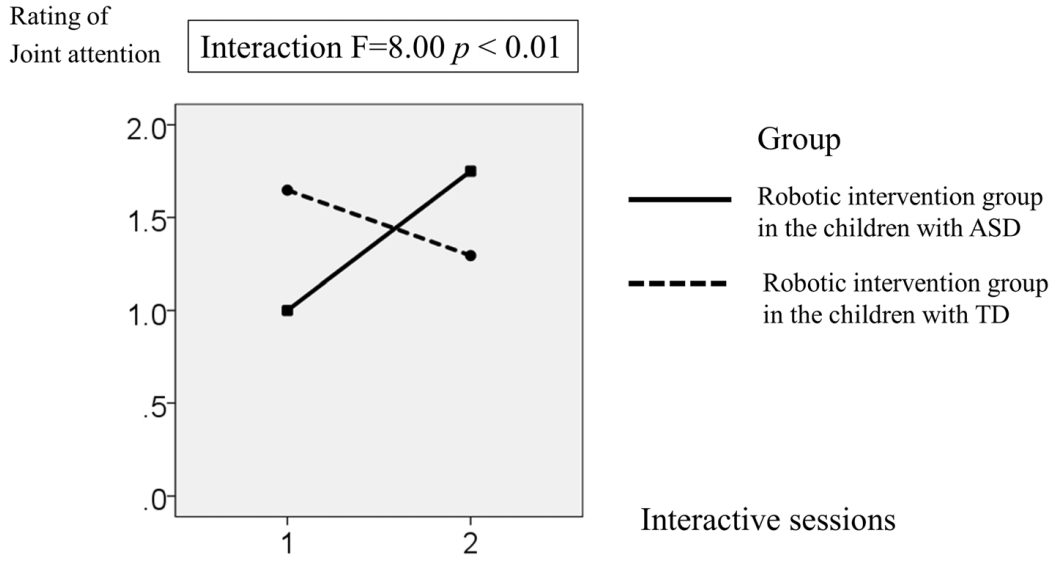}}
    \caption{Two-way mixed ANOVA with one repeated factor (time, i.e., the first and second sessions) and one group factor (robot group vs. control group) with $\alpha = 0.05$.}
    \label{fig:im6}
\end{figure}

\begin{figure}[!htbp]
    \centering
    {\includegraphics[width=1\linewidth]{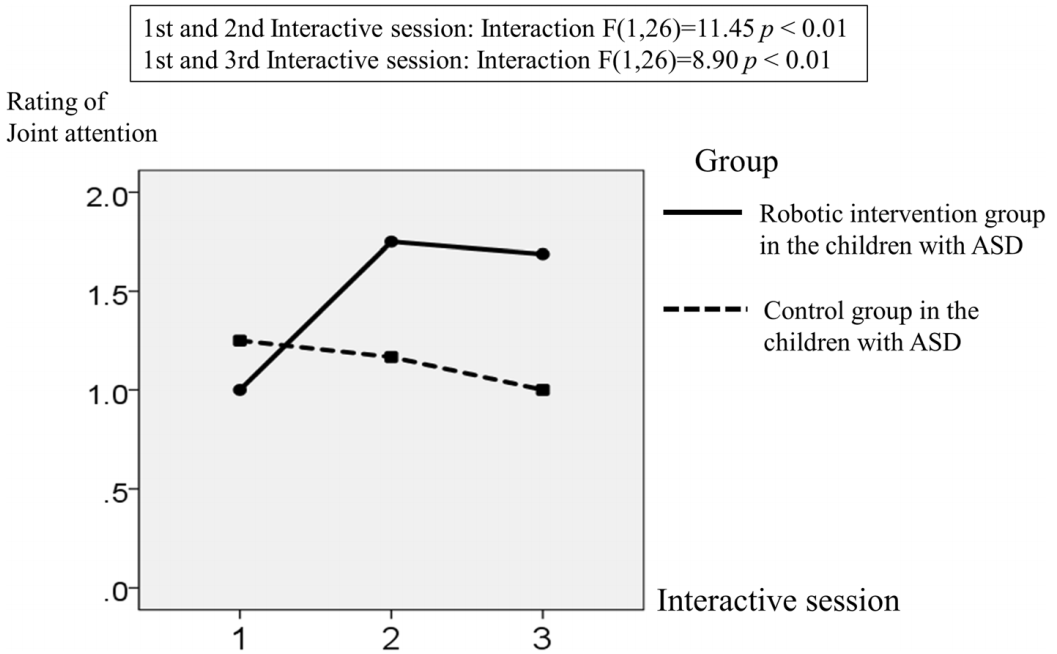}}
    \caption{Two-way mixed ANOVA with one repeated factor (time, i.e., the first and third sessions) and one group factor (robot group vs. control group) with $\alpha = 0.05$.}
    \label{fig:im7}
\end{figure}

Concerning the differences in the ratings of joint attention performance between the robotic interaction and human agent groups in the autistic children, it was employed a two-way mixed ANOVA with one repeated factor (time, i.e., the first and second sessions) and one group factor (robot group vs. control group) with an alpha factor equal to 0.05. 
This analysis of the variance revealed a significant interaction between the time and group effect, as depicted in figure \ref{fig:im6}. This result confirmed the first research group's hypothesis, i.e., that autistic children would exhibit better joint attention capabilities under the CommU condition than under the human agent condition.
%%%%%%%%%%%%%%%%%%%%%%%%%%%%%%%%%
%%%%%%%%%%%%%%%%%%%%%%%%%%%%%%%%%
\newline
Moreover, a slightly modified two-way mixed ANOVA was also applied with one repeated factor (time, i.e., the first and third sessions) and one group factor (robot group vs. control group) with an alpha factor equal to 0.05 (see figure \ref{fig:im7}). The results of this latter analysis supported the second hypothesis advanced by the research team, i.e., that children with autism spectrum disorders would present advancement in joint attention tasks in human-human interactions after interacting with CommU. \newline Concerning the differences in the ratings of the joint attention under the robotic condition between the autistic children and their typically developed counterparts, the analysis' results plotted in figure \ref{fig:im7} confirm the third hypothesis advanced by the research group. Precisely, the two-way mixed ANOVA results with one repeated factor (time, i.e., first and second sessions) and one group factor (i.e., autistic group vs. typically developed group) revealed a meaningful interaction between the group effect and the time. This result confirmed the research team's third hypothesis, i.e., the improvements in joint attention tasks owing to robot intervention would be more significant in children with autism spectrum disorders than in typically developed children, in light of their non-social preferences.
%%%%%%%%%%%%%%%%%%%%%%%%%%%%%%%%%
%%%%%%%%%%%%%%%%%%%%%%%%%%%%%%%%%
Collectively, the obtained results suggest that the robot-mediated intervention has successfully improved joint attention capabilities in autistic children. Consequently, it can be stated that the performance improvement is clinically relevant. Specifically, the results have demonstrated that autistic children exhibited better joint attention during their interaction with the robot, which has clear eyes and can turn them, than when interacting with the human agents. Moreover, the results show also that children with autism spectrum disorders presented improved joint attention tasks with a human agent after the interaction with the robot. In other terms, the acquired abilities generalize to human-human interactions. For these reasons, it could be stated that the usage of robots as adjunctive tools for short-term training in autistic individuals is recommended. 
\subsection{Limitations}
This subsection intends to acknowledge the most relevant methodological limitations presented by this research. 
\begin{itemize}
    \item \textit{Intervention's limited duration} The most impactful limitation is the research's limited duration. Specifically, the study consisted of a single session that did not determine whether the children responded similarly over multiple sessions. In fact, to obtain a broader understanding of the habituation to the robotic agent over time, it would have been necessary to observe the acquired joint attention capabilities over a more extended period.
    \item \textit{Intervention's effect on the long-term horizon} The ultimate purpose of the interventions is to improve children's communication abilities in daily-life. Although there are promising results that suggest that the communication improvements in robot-human interaction generalize to human-human interactions, there is no evidence supporting the generalizability of acquired joint attention capabilities to daily-life (e.g., at home) due to the short-term horizon considered by the research's study.
    \item \textit{Applicability in clinical scenarios} Another limitation is constituted by the fact that the participants involved in the study had average cognitive abilities. In order to evaluate the protocol's effectiveness in real clinical scenarios, it would be necessary to integrate a higher number of individuals with a broader range of autistic traits into the population sample. In doing so, it would be possible to fully understand robotic interventions' potential application and impact in clinical circumstances.
\end{itemize}
\section{Comparative reading: choosing the optimal robot and the need for structured interactions}
As previously stated in subsection \ref{txt:proposedApproach}, when two papers exhibit a high degree of similarity in terms of their underlying logic, we provide a comparative reading that focuses on how a particular work overcomes the limitation of the other one jointly with a critical analysis of their common limitations. In this case, the comparison covers the previously presented papers.
\subsubsection{The Effectiveness of Using a Physical Robot} Several interventions employ non-social objects for autistic children. For instance, virtual reality usage allows the construction of multiple scenarios by limiting the required resources. However, humanoid robot-assisted interventions could be more attractive to children than two-dimensional programs thanks to a robot's physical presence, resulting in more engaging and enjoyable interactions (\cite{lee2006physically},\cite{pan2016comparison}). Considering the importance of active participation in joint attention tasks, the success of the intervention strongly depends on the fact that the participant show sustained motivation to interact with the robot.
\subsubsection{The optimal choice of the robot}
The choice of the robot plays a crucial role in the intervention's effectiveness. For example, several research studies have demonstrated that autistic children exhibited enhanced performance in imitation tasks after a robot-mediated intervention. However, the choice of the most suitable robot is strictly dependent on the task. Multiple research studies (\cite{anzalone2014children},\cite{warren2015can},\cite{bekele2014pilot}) have proved this aspect while investigating the robot-mediated intervention efficacy in joint attention tasks on autistic children. Precisely, these studies have reported unsatisfactory results due to the usage of a robot that could not turn its eyes (i.e., "Nao"). As observed by Warren et al. \cite{warren2015can} and Bekele et al. \cite{bekele2014pilot}, eye-gazing is a fundamental aspect of joint attention tasks, and the robot's incapacity to turn its eyes prevented it from resembling a human agent. Moreover, this limitation must be added to the ones previously described in section \ref{txt:techlimit1}). On the contrary, the social robot CommU is particularly suitable for joint attention tasks. In fact, it features a wide range of gaze expressions thanks to the thoughtful design of the eyes and many degrees of freedom dedicated to controlling its vision field. The robot can realize a wide range of simplified expressions that can help autistic children to focus their attention on the most relevant traits for joint attention tasks. In addition, CommU resembles a child, and thus, it is easy to anthropomorphize. Another benefit of employing this robot for joint attention tasks is that its cute appearance is expected to help prevent fearfulness among children. Finally, in contrast to the Nao robot, CommU makes very little noise when moving, preventing the distress of children who are particularly sensitive to loud noises, which is a common trait in autistic children. As Madipakkam et al. have observed \cite{madipakkam2017unconscious}, autistic children manifest an atypical response to eye contact since they tend to unconsciously avoid it. In this respect, CommU's clear eyes are likely to stimulate autistic children to note their existence. Previous studies have confirmed this point using the Flobi robot, which has visible eyes and can turn them. These characteristics pushed autistic children to note the robot's eyes \cite{damm2013different}. In contrast, in previous research using Nao, whose eyes are relatively small, although the autistic children's attention seemed to be captured by the robot, they could not focus on its eyes (\cite{anzalone2014children},\cite{warren2015can},\cite{bekele2014pilot}). It is common knowledge that  Nao is a powerful attractor for children with autism spectrum disorders. In fact, Nao does not profoundly resemble a human; hence autistic children do not feel any pressure. However, Nao's brightly colored body parts are likely to prevent the children from focusing on relevant social signals (i.e., the robot's eyes) \cite{pennisi2016autism}. In fact, the robot's body components must not be so bright as to over-stimulate the child to prompt joint attention. On the contrary, the discreet color of CommU's body parts facilitates joint attention tasks. \hfill \break \indent Pierno et al. \cite{pierno2008robotic} confirmed that facilitation effects in joint attention tasks were noticed solely under the human agent condition in typically developed children and only under the robot condition in autistic children. This result is effectively explainable because the tasks' complexity completed by the robot differs considerably from the tasks' complexity executed by its human counterpart. This is because human behavior includes the body's and head's pose and rotation, facial expression, and unique unintentional gestures not present in the robot's motions. In other terms, there are many potential uncontrolled variables in the experiments with the human partner that prevent autistic children from achieving high performances in joint attention tasks. Note that the same outcome would be obtained by employing a more complex robot like iCub \cite{metta2010icub}, which has an extremely high number of potential uncontrolled variables.
\subsubsection{Partially Structured Interactions}
As previously pointed out in section \ref{txt:structuredInt_general}, designing profoundly structured interactions is a crucial aspect to build structured scenarios in which particular social behaviors may occur, finally maximizing interventions' effectiveness. The present critical reading brings out a characteristic common to the previously presented research studies: both the interventions' interactions are only partially structured. 
In fact, although the proposed interaction by Kumazaki et al. \cite{kumazaki2018impact} has a clear goal (i.e., promoting children's joint attention towards the target), its structure is very similar to a free-play scenario. In other terms, except for the last prompts, the robot's actions are mainly directed towards the children's entertainment instead of focusing on stimulating specific social behaviors. This limitation has been overcome by Warren et al. \cite{warren2015can}, which shows a more well-thought-out structure of prompts that varies according to the participants' responses. In fact, as described in section \ref{txt:structuredInt_paper1}, the interaction development depends on a well-defined system of prompts levels and rewards whose goal is to encourage the expected behaviors. 
\hfill \break \indent 
Although the latter structure could be considered adequate, a further step can be carried out. A crucial issue is that it is complicated to compare different works and even measure the approach's effectiveness if the proposed interactions are not rigorously structured. This is because the interactions' structure will be profoundly dependent on the task to be executed. However, it is possible to solve this problem by developing a rigorous mathematical model regulating prompts and reward systems. In other words, this will constitute the system's underpinning the interaction's structure that will be subsequently adapted dependently on the requirements, desired contents, and clinical goals. In doing so, it is possible to overcome the limitation of an inadequately structured interaction while retaining the versatility required to adapt the model to different scenarios.
\section{Overcoming the Absence of a Formal Model and the Presence of a Human in the Control Loop}
\label{txt:paper3sect}
\subsection{Introduction}
As observed in the preceding sections, several previous robot-mediated interaction experiments were based on free interactions. However, recent studies have proved that robot-mediated interventions are more effective when focused on the core impairments of autism spectrum disorder (e.g., joint attention deficit). Moreover, most of these early human-robot interaction systems were open-loop systems, not capable of adapting and individualizing interventions because they were not responsive to the dynamic interaction feedbacks from the participants. However, tackling core deficits using robot-mediated intervention is more complicated since the autonomous system needs to elicit responses in the sphere of the core deficit through a set of well-designed interaction protocols, evaluate participant's response in real-time, and adapt the robot's own interaction according to the formal model.
\hfill \break \indent
Zheng et al. \cite{zheng2017design} have introduced an innovative joint attention intervention system for autistic that overcomes several limitations presented in the previous dedicated sections. Specifically, the proposed system aims to overcome the need to employ body-worn sensors, non-autonomous robot operation demanding human involvement, and, above all, the absence of a formal model regulating prompts level and rewards. Precisely, the research group has proposed an entirely autonomous robotic system, called a non-contact responsive robot-mediated intervention system, that can infer attention through a distributed non-contact gaze inference mechanism. Concerning the formal model, the system employs an embedded least-to-most (LTM) robot-mediated interaction model.
\subsection{Employed system's structure}
The scheme proposed by the research team provides an entirely autonomous closed-loop interaction between the participant and the system, whose constitute components are enlisted below:
\begin{itemize}
    \item \textit{Robot module} The robot employed for the research study was the humanoid robot NAO, widely applied for children with autism spectrum disorders due to its attractive childlike appearance and high controllability. Further, the research group has developed a novel controller for NAO that communicates with the supervisory controller.
    \item \textit{Target module} Two target monitors located to the participant's left and right sides were used as attentional targets. Both monitors were commanded independently by two target controllers that received commands from the supervisory controller. Expressly, the robot would point to one of the monitors at a time and ask the participant to look at what was being displayed on that monitor.
    \item \textit{Gaze-tracking module} The gaze-tracking module identified the participants’ gaze behavior. Specifically, the direction of each participant’s gaze was computed through the orientation of his head, which was detected by a set of cameras.
    \item \textit{Supervisory Controller} The supervisory controller had the role of communicating with various system components and controlling the interaction's global logic.
\end{itemize}
%%%%%%%%%%%%%%%%%%%%%%%%%%%%%%%%%
\begin{figure}[!htbp]
    \centering
    {\includegraphics[width=1\linewidth]{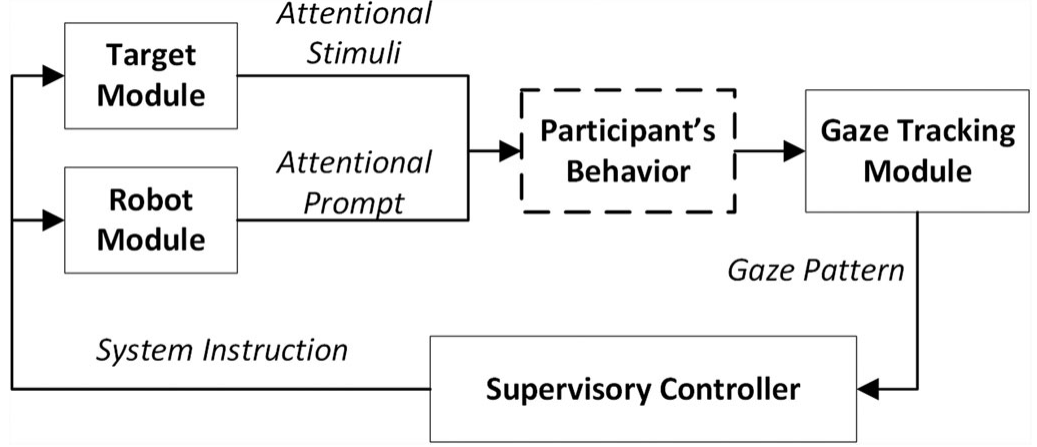}}
    \caption{Architecture of the employed system.}
    \label{fig:im8}
\end{figure}
%%%%%%%%%%%%%%%%%%%%%%%%%%%%%%%%%
\subsection{Trial Characteristics}
The human-robot interaction using the proposed system works as follows. An autistic child is positioned in front of the humanoid robot. The room is equipped with a set of spatially distributed target monitors where audiovisual stimuli are presented. The robot administers a LTM-based joint attention prompting protocol to the child, and a set of distributed cameras will infer the child’s response in terms of gaze direction. Depending on whether the interaction ends with a target hit (i.e., the child shares attention or not), the robot provides appropriate feedback and proceeds to the following prompt, according to the protocol.

\subsection{Prompts and Reward Provision System: The LTM Interaction Protocol}
\label{txt:LTM_def}
\subsubsection{The Protocol's Structure}
In recent years, the LTM protocol has been widely applied in diagnostic and screening tools for children with ASD (\cite{lord2000autism},\cite{mundy1996preliminary}). The LTM interaction protocol consists of an underlying logic that is not limited to a specific intervention but can be applied as a general guidance mechanism for robot-mediated therapies. Specifically, the research team has applied the LTM hierarchy to generate the interaction protocol. In LTM, the teacher leaves the learner the opportunity of responding autonomously on each training stage and provides the least intrusive prompt first. If needed, more intrusive prompts, generally upgraded based on the previous prompts, are provided to the learner to complete each training procedure \cite{libby2008comparison}. For the sake of clarity, we provide the reader the most relevant examples of its application to robot-mediated intervention systems for autistic children.
\begin{itemize}
    \item Kim et al. \cite{kim2014designing} devised a robot-assisted pivotal response training system, where the higher levels of prompts were built by adding target responses suggestions on lower level of prompts.
    \item Huskens et al. \cite{huskens2013promoting} employed a robot to prompt question-asking behaviors in autistic children. The robot used an open-question prompt initially. If the participants did not answer correctly, the robot would add more hints (e.g., providing a portion of the correct response) in the following prompts. 
    \item Esubalew et al. \cite{esubalew2012step} proposed the ARIA system to teach joint attention skills to children with autism spectrum disorders. If the participants did not respond to the robot's basic directional prompts, more intrusive prompts with additional visual and verbal directional hints were given.
    \item Feil-Seifer and Mataric \cite{feil2012simon} and Greczek et al. \cite{greczek2014graded} proposed a graded cueing feedback mechanism whose goal was to teach imitation skills to children with autism spectrum disorders. Specifically, higher prompts were upgraded based on the initial prompt by adding further verbal and gestural cues to assist children in copying gestures. 
    \item Zheng et al. \cite{zheng2015robot} proposed a robot-mediated imitation learning system utilizing a prompting protocol to address incorrect imitations. Specifically, the robot first exhibited a gesture, asking the participant to imitate it. If the child could not perform it correctly, the robot would suggest how to improve it in the following prompts.
\end{itemize}
\subsubsection{Contribution to the State of the Art}
Although the application of social robots to design interventions for children affected by ASD has unquestionably captured the research community's interest, no mathematical model has been formalized to design a general LTM framework before the one introduced by the paper considered in this analysis. The research team has attempted, for the first time, to define a general LTM-based robot-mediated intervention (LTM-RI) model. Specifically, this model can be employed to facilitate the development of different abilities in autistic children and adjust the prompts for a specific skill or scenario. The following paragraphs aim to analyze in detail the structure and the logic underlying the formal model. Secondly, we briefly describe how the model is adapted into the presented use case (i.e., a joint attention intervention).
\subsubsection{The Model's Foundations}
Generally, intervention systems employ prompts to train autistic children in specific actions. These prompts consist of robot actions (e.g., motions, gestures, and speeches) and often include environmental factors that correlate with the robot’s actions (e.g., the attentional target the robot has to refer to in joint attention tasks).\newline
To provide the reader a concise but comprehensive overview of the procedure defined by the formal model, we first define two sets of objects, denoted robot actions \{$RA$\} and environmental factors \{$EF$\}, such that it is possible to combine different $RA_i$ and $EF_j$ to generate various prompts. The results of these combinations will differ in terms of the probability of eliciting the expected response $ExpResp$, from a child (e.g., the child looking at the target monitor.) For instance, in the current joint attention study case, the robot turning its head ($RA$) to a static picture displayed on the target monitor ($EF$) is very likely to have a weaker influence on the children than turning its head while pointing ($RA$) to a cartoon video displayed on the target monitor ($EF$). \newline From the theoretical point of view, it is possible to define order relationships between objects belonging to the same set. For instance, it would be beneficial for the present study case to order the elements in the sets \{$RA$\} and \{$EF$\} by employing the specially defined relationship \textit{Stronger(.)} (i.e., includes more informative content). This relation can be applied to both the sets and works as follows:
\begin{equation*}
Stronger(Ra_a,Ra_b) \triangleq 
\left\{
\begin{alignedat}{2}
 True\;\;\;\;\;\;\;\;\;\; \text{if}\;a>b\\
 False\;\;\;\;\;\;\; \text{otherwise}\\
\end{alignedat}
\right.
\end{equation*}
\begin{equation*}
Stronger(EF_c,EF_d) \triangleq 
\left\{
\begin{alignedat}{2}
 True\;\;\;\;\;\;\;\;\;\; \text{if}\;c>d\\
 False\;\;\;\;\;\;\; \text{otherwise}\\
\end{alignedat}
\right.
\end{equation*}
Please note that the critical notion underpinning the relationship \textit{Stronger(x,y)} is the one that determines whether the argument $x$ is greater than the argument $y$ (i.e., that a particular stimulus has a more significant impact on the child than another one). Unfortunately, to determine these notions' actual correctness, we can only rely on common sense and clinical experiences.
\subsubsection{From the formal model to the algorithm}
\label{LTM_RI_formal}
LTM-RI begins by providing the weakest combination of $RA$ and $EF$ to generate the least intrusive prompt. If this cannot elicit the child's expected response (\textit{ExpResp}), stronger $RAs$ and $EFs$ will be employed iteratively to build more instructive prompts until the intervention terminates. The iterative procedure presented above can be formalized as follows.
%%%%%%%%%%%%%%%%%%%%%%%%%%%%%%%%%
\begin{figure}[!htbp]
    \centering
    {\includegraphics[width=1\linewidth]{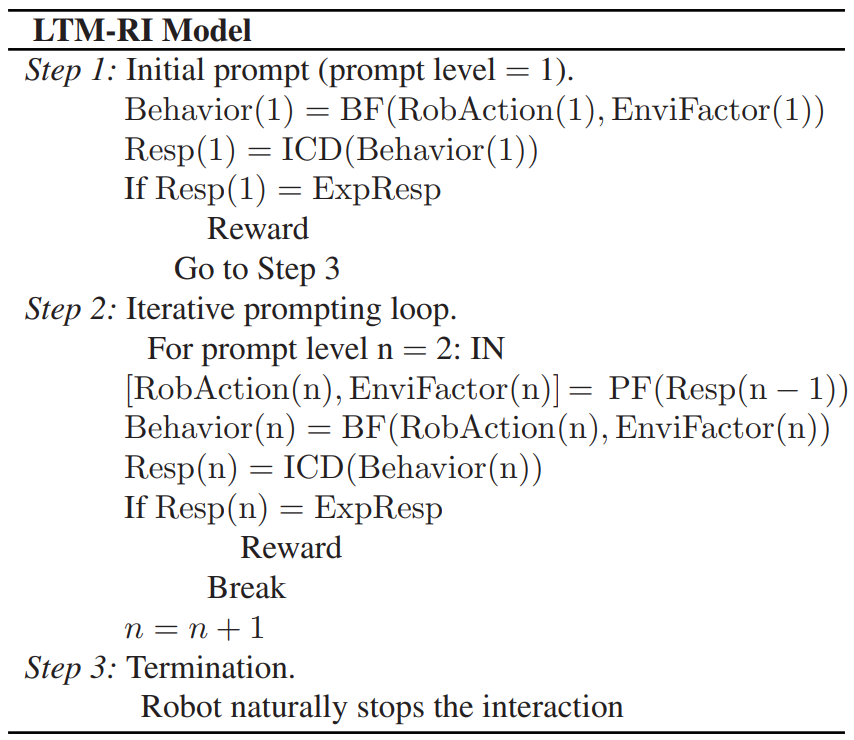}}
    \caption{Formalization of the LTM-RI model.}
    \label{fig:im9}
\end{figure}
%%%%%%%%%%%%%%%%%%%%%%%%%%%%%%%%%
\newline
With reference to figure \ref{fig:im9}, we highlight the following points:
\begin{itemize}
    \item BF is an implicit function that takes as inputs a \textit{RA (RobAction)} and an \textit{EF (EnviFactor)} and returns the representation of the participant's behavior (e.g., participant's gaze direction).
    \item The interactive cue detection function \textit{(ICD)} takes as input the participant's behavior and determines whether the behavior is \textit{ExpResp}. The implementation of such function profoundly varies depending on the system's design choice and clinical goals. For instance, in the current study, \textit{ICD} is the gaze-tracking module. Generally, behaviors exist on a spectrum; hence it is particularly complex to determine if a response is the expected one. However, in the simplest scenario, it is possible to categorize \textit{Resp(n) $=$ ExpResp} and Resp(n) $\neq$ ExpResp.
    \item PF is the prompting function determining which \textit{RA} and \textit{EF} to combine and provide to the participant if \textit{Resp(n) $\neq$ ExpResp}. At the implementation level, \textit{PF} is a sorted list of prompt levels following the LTM hierarchy. As previously observed, it is necessary to employ the least intrusive prompt required by the participant to perform \textit{ExpResp}. This latter procedure is implemented as follows. If \textit{Resp(n) $\neq$ ExpResp} (i.e., the response detected is not the expected response), given \textit{RobAction$(n-1)=RA_i$} (i.e., robot action at time instance $n-1$), and \textit{EnviFactor$(n-1)=EF_j$} (i.e., environmental factor at time instance $n-1$), we extract the robot action for the next instant, \textit{RobAction(n)$=RA_l(l \geq i)$} , from \textit{\{RA\}} and the environmental factor for the next time instance, \textit{EnviFactor(n)$=EF_k (k \geq j)$}, from \textit{\{EF\}}, so that the next prompt repeats the last prompt or provides a new prompt with greater informative content.
\end{itemize}
\subsubsection{Algorithm's breakdown}
This subsection aims to provide the reader a concise description of the algorithm's steps.
\begin{enumerate}[i)]
    \item \textit{Step 1:} In Step 1, the participant’s baseline behavior \textit{Perf(1)} is assessed by prompt level 1, which consists of the weakest combination of \textit{RA (i.e., RobAction(1))} and \textit{EF (EnviFactor(1))}. If the participant’s response, \textit{Resp(1)}, is the expected one \textit{(i.e., Resp(1) = ExpResp)}, then higher prompts are unnecessary. Hence, the system provides the participant the reward and then proceeds to Step 3 to conclude the intervention. Otherwise, Step 2 is executed.
    \item \textit{Step 2:} In Step 2, the algorithm starts an iterative prompting loop whose goal is to provide increasingly intrusive prompts until the participant's expected response is detected.  The iterative loop starts with the prompt level 2. If the participant does perform the expected response \textit{(ExpResp)}, the higher prompts are given until the participant provides the expected response or level IN is reached. During the iterative loop, the next level of prompt \textit{(RobAction(n)} and \textit{EnviFactor(n))} is generated based on the current response of the participant \textit{Resp($n-1$)}, according to the \textit{PF} function previously defined. If \textit{Resp(n)$=$ExpResp}, the system gives a reward to the participant and proceeds to Step 3.
    \item \textit{Step 3:} The robot naturally terminate the interaction.
\end{enumerate}
Referring to the previous analysis, it is useful to investigate the relationship between the iterative procedure's probability of eliciting the expected response and the prompts' level. From the algorithm's structure follows that if ExpResp has been detected only on prompt level $n$, it means that the previous levels, from 1 to $n-1$, have been executed but failed to elicit $ExpResp$. At this point, we proceed in building a generalization of the proposed model. \newline Suppose $ER_n$ means $Resp(n)=ExpResp$, and $PTx$ denotes the prompt level $x$ had been executed but was unsuccessful. Then, $P(ER_n\vert PT_1, \dots ,PT_{n-1})$ will represents the probability that $ExpResp$ happens on prompt level $n$. Hence, to measure the impact of LTM-RI, we define an intensity function $I_n$ as follows:
\label{txt:IntensityFunction}
\begin{align*}
& I_n \triangleq P(ER_1)\;+\;P(ER_2 \vert P(ER_1))\;+ \\
& + \;\dots\;+\\ 
& P(ER_n \vert P(ER_1) + \dots + P(ER_{n-1})), \\ 
\end{align*}
where $I_n$ denotes the probability of $ExpResp$ at or before prompt level $n$. Hence, we observe that, from the theoretical point of view, we can virtually extend the formal model to the desired length. In this respect, we note that the LTM-RI procedure has two goals:
\begin{enumerate}
    \item \textit{Goal 1:} Given $m > n$, $I_m > I_n$, i.e., the addition of prompt levels increases the probability of $ExpResp$.
    \item \textit{Goal 2:} $I_{IN}=1-\epsilon$, where $\epsilon=0$ or $\epsilon$ is a small positive number. In other terms, at the highest prompt level, the system can elicit $ExpResp$ with a very high probability.
\end{enumerate}
However, in practice, it is unquestionably necessary to determine the correct trade-off between clinical goals and interaction length to avoid distressing the participants with excessively long interventions. 
\subsubsection{Applying LTM-RI to the current study case}
\begin{figure}[!htbp]
    \centering
    {\includegraphics[width=1\linewidth]{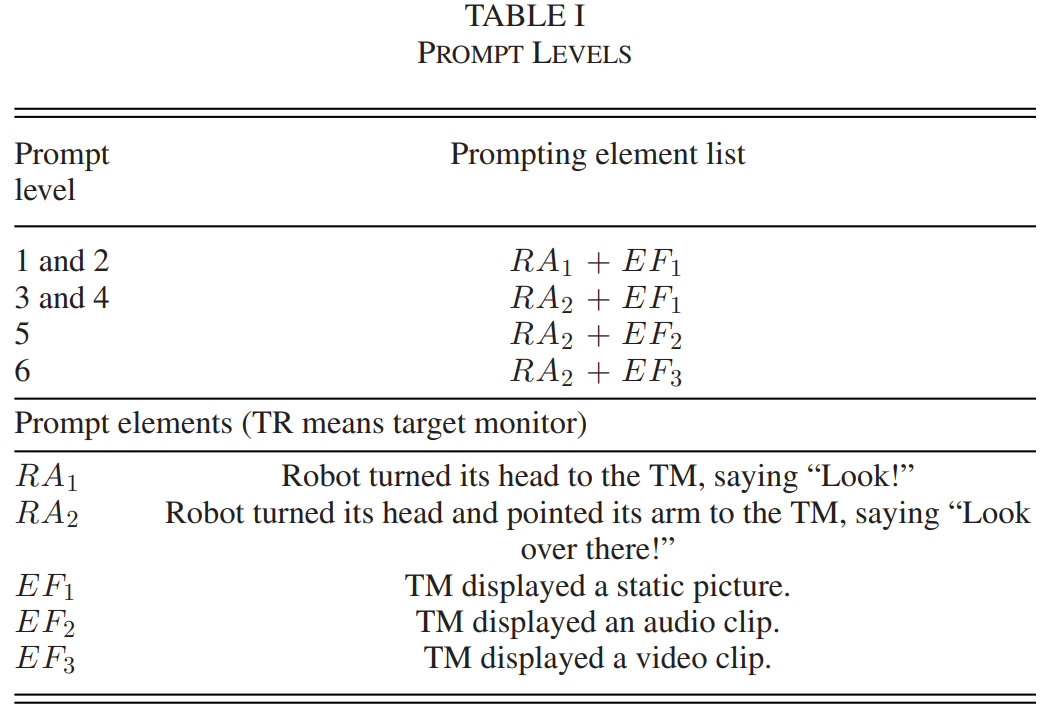}}
    \caption{An instance of the formal model applied to a joint attention task.}
    \label{fig:im10}
\end{figure}

As previously pointed out, LTM-RI is a general model not restricted to a particular skill. In fact, the participant's behaviors tracked by the model depend on the ICD function's design. Similarly, the number of prompt levels and their content can be easily adapted within this framework by changing the implementation of the PF function. Concerning the implementation proposed by the research team for the proposed study case, the specific content of the prompts in the six-level hierarchy was inherited by previous studies to demonstrate the model's versatility. Figure \ref{fig:im10} describes the PF function in the LTM-RI model, which denotes how RAs and EFs are combined. For the sake of conciseness, we denote more intrusive directional information with larger subscripts. For example, we present a sample interaction extracted from the current study case.
%%%%%%%%%%%%%%%%%%%%%%%%%%%%%%%%%
%%%%%%%%%%%%%%%%%%%%%%%%%%%%%%%%%
\newline
According to the algorithm's breakdown previously presented, we briefly analyze the prompts' levels' hierarchy:
\begin{itemize}
    \item \textit{LTM-RI step 1 (prompt level 1):} The robot turned its head to the target monitor, saying “Look!” ($RA_1$). At the same time, the monitor displayed a static picture ($EF_1$).
    \item \textit{LTM-RI step 2, where $IN=6$:}
        \begin{itemize}
            \item Prompt level 2: the prompt was the same of prompt level 1.
            \item Prompt level 3-4: the robot not only turned its head, but also pointed its arm to the target monitor, saying "Look over there!" ($RA_2$). At the same time, the monitor still displayed a static picture ($EF_1$).
            \item Prompt level 5-6: the robot action was identical to $RA_2$, but the monitor displayed an audio clip ($EF_2$) and a video clip ($EF_3$), respectively. At any time during a trial, if the participant looked at the target (i.e., $Resp=ExpResp$), the robot would say \textit{"Good job!"} and the target monitor would display a cartoon video for 10 seconds as rewards. 
        \end{itemize}
    \item \textit{Prompt level 6 - Termination:} If the expected response was not detected, the prompt level would be proposed one by one until the completion of prompt level 6. In this latter case, the intervention terminates regardless of the interaction's result.
\end{itemize}
\subsection{Results}
Concerning the obtained results, the research group has measured the system's effectiveness according to two distinct criteria: preferential attention and joint attention performance, defined below:
\begin{itemize}
    \item \textit{Preferential Attention:} firstly, the research team has evaluated the participants' attention on the robot and target monitors to measure the participants' engagement. Precisely, for each object, it was defined a region of interest that covered that object with a margin around it. Then it has been analyzed how their attention was distributed among the robot and the target monitors. \newline The hypotheses the research team aimed to verify by computing this metric are the following:
    \begin{enumerate}
        \item The participant would have paid significant attention to the robot because the robot was the primary interactive agent.
        \item The participant would have paid more attention to the target monitor than the non-target monitor because the target monitor was referred to by the robot and displayed visual stimuli.
        \item If the participants' interest in the robot was sustained over the sessions, their time spent looking at it would not change significantly.
    \end{enumerate}
    \item \textit{Joint attention performance:} secondly, the authors have evaluated the participants' joint attention performance that basically reflects the system's effectiveness. More precisely, for each session, they have computed:
    \begin{itemize}
        \item The number of trials in which the participants hit the target successfully.
        \item The average prompt levels the participants needed to hit the target.
    \end{itemize}
    \begin{enumerate}
    \setItemnumber{4}
        \item If the robotic intervention was effective, the participant's performance in joint attention tasks would have improved significantly.
        \item If the participants' joint attention skills have improved, they would have observed higher intensity values in low-prompt levels than in high-prompt levels.\footnote{With intensity, we are referring to the Intensity function $I_n$ previously defined in section \ref{txt:IntensityFunction}.}
    \end{enumerate}
\end{itemize}
%%%%%%%%%%%%%%%%%%%%%%%%%%%%%%%%%
\begin{figure}[!htbp]
    \centering
    {\includegraphics[width=1\linewidth]{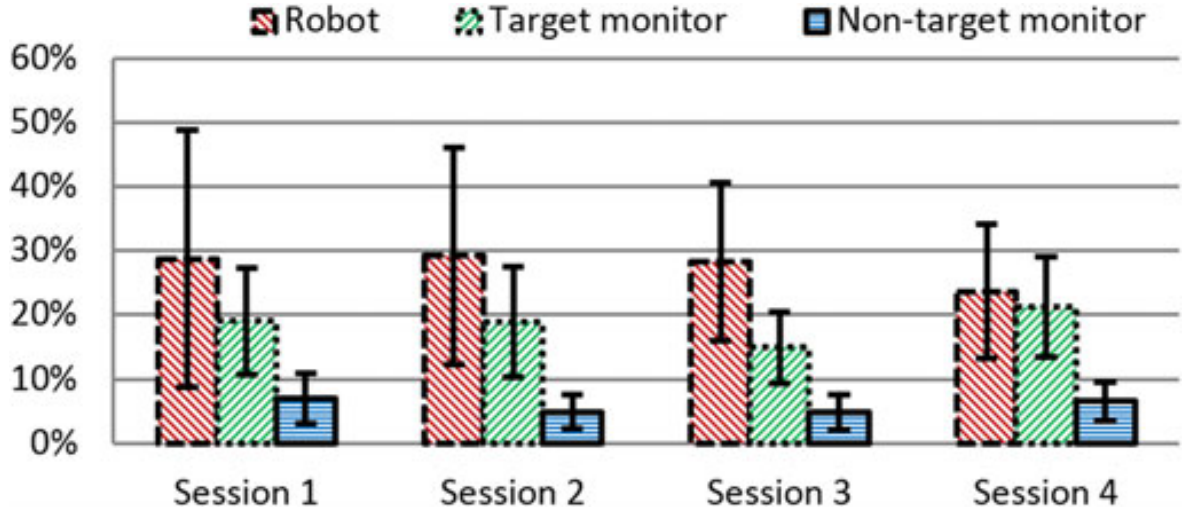}}
    \caption{Percentage of the session time participants spent looking at the robot, target monitor, and the non-target monitor.}
    \label{fig:im11}
\end{figure}
%%%%%%%%%%%%%%%%%%%%%%%%%%%%%%%%%
Figure \ref{fig:im11} shows the percentage of the session time participants spent looking at the robot, target monitor, and the non-target monitor. The obtained results validate the first three hypotheses. In fact, from the figure, it is possible to make the following inferences:
\begin{itemize}
    \item Hypothesis 1) $\rightarrow$ The participants looked at the robot more than the monitors in every session.
    \item Hypothesis 2) $\rightarrow$ The participants looked at the target monitor much more than the non-target monitor. This phenomenon is probably due to the following reasons. Firstly, the robot referred to the target monitor, and the participants responded more to the referred direction. Secondly, the target monitor displayed visual stimuli during prompts and rewards, which caught and held the participants' attention.
    \item Hypothesis 3) $\rightarrow$ The research group has compared the time that the participants spent looking at the robot across all sessions and found no statistically significant change (i.e., $p = 0.9515-0.1937$). This result suggests that the participant's interest in the robot was held throughout the sessions, thus validating the third hypothesis.
\end{itemize}
It should, however, be mentioned that due to the heterogeneous neurodevelopmental trajectory and behavioral pattern of children affected by autism spectrum disorders, the participants exhibited considerably different attention patterns and joint attention skills. Consequently, figure \ref{fig:im11} presents large standard deviations in all cases. On the contrary, the standard deviation of the looking time on the robot lowered from sessions 1 to 4, showing that the participants' looking toward the robot tended to be stable after a few intervention sessions.
%%%%%%%%%%%%%%%%%%%%%%%%%%%%%%%%%
\begin{figure}[!htbp]
    \centering
    {\includegraphics[width=1\linewidth]{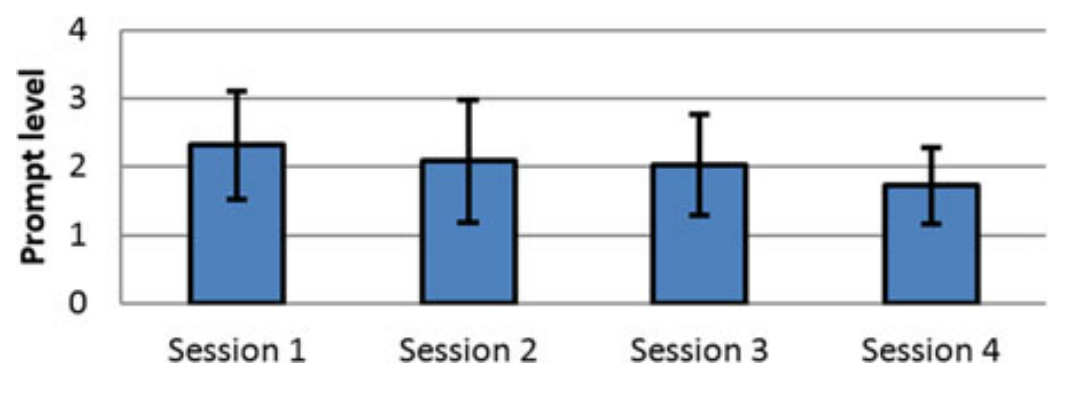}}
    \caption{Average target hit prompt levels in all sessions.}
    \label{fig:im12}
\end{figure}
%%%%%%%%%%%%%%%%%%%%%%%%%%%%%%%%%
\newline
Observing the results above enables us to validate the last two hypotheses. Figure \ref{fig:im12} shows the average target hit prompt levels in all sessions. Note that the lower the prompt level needed by participants, the better their performance was. The authors noted that from sessions 1 to 4, the average target hit prompt level decreased monotonically from 2.31 to 1.71. Moreover, a Wilcoxon-signed rank test showed that the decrease in prompt level from sessions 1 to 4 was statistically significant ($p = 0.0115$), thus indicating that the participants’ performance improved significantly and confirming the fourth hypothesis.
%%%%%%%%%%%%%%%%%%%%%%%%%%%%%%%%%
\begin{figure}[!htbp]
    \centering
    {\includegraphics[width=1\linewidth]{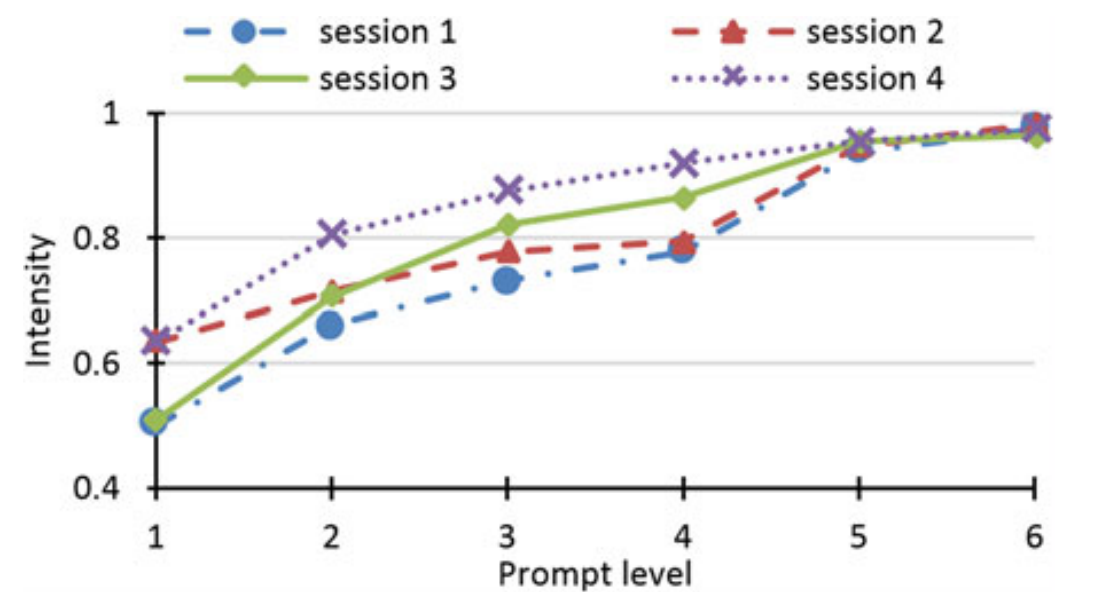}}
    \caption{Intensity of prompt levels across four sessions.}
    \label{fig:im13}
\end{figure}
%%%%%%%%%%%%%%%%%%%%%%%%%%%%%%%%%
\newline
Moreover, the authors have evaluated how the incremented LTM prompt levels elicited target hit behavior, i.e., whether the two goals discussed in section \ref{txt:IntensityFunction} were achieved. Precisely, the computation is performed as follows:
\begin{align*}
& I_n = P(ER_n \vert P(ER_1) + \dots + P(ER_{n-1})) =\\ 
& = \frac{\text{Number of trials ended with target hit on prompt level n}}{\text{Total number of trials}}
\end{align*}
For the sake of conciseness, we denote the intensity of prompt level $n$ in session $x$ as $I^{x}_n$. Figure \ref{fig:im13} depicts the values of $I$ in sessions 1–4. With reference to it, the research group has moved the following observations:
\begin{itemize}
    \item Given $a>b$, $I^{x}_a > I^{x}_a$ in all the sessions sessions, thus indicating that adding more instructive prompt levels on top of low prompt level elicited more target hits. Therefore, Goal 1 was achieved.
    \item Given $x>y$, figure \ref{fig:im13} also shows that for the same prompt level, $I^{x}_a > I^{y}_a$ or $I^{x}_a \approx I^{y}_a$ in most of the cases. Please note there is an exception: $I^{2}_1 > I^{3}_1$. Hence it is possible to infer that, in general, as the participants had received more interventions in later sessions, their possibilities of a target hit on the same prompt level grew with occasional fluctuations. Note that at the end of the clinical trial $I^{1}_6$ to $I^{4}_6$ were 0.97, 0.98, 0.96, and 0.97, respectively. Hence it can be concluded that the LTM-RI trial could eventually support participants to hit the target in almost all of the trials. Thus, the prompt level content and the number of the levels ($IN = 6$) were adequately designed for the proposed task. Therefore, Goal 2 was also achieved.
\end{itemize} 
\subsection{Limitations}
This subsection aims to acknowledge the most relevant methodological limitations presented of the employed approach.
\begin{itemize}
    \item \textit{Limited population sample:} as usual for clinical trials, the research group could not validate the system on a large population sample. Although the proposed system's results are particularly promising, it needs to be tested in formal clinical studies in the future. 
    \item \textit{Absence of a systematic comparison with different methods:} the research team has assessed promising joint attention capabilities within the system. However, the presented research study lacks a comprehensive and systematic comparison with different methods.
    \item \textit{Generalization of the acquired skills:} the presented research study has registered a considerable improvement of the joint attention skills in the participants. However, it still lacks an investigation regarding if such training can be generalized to human-human interactions. An idea to address this issue consists of conducting pre-and post-tests (e.g., interactions' video recording) to observe how joint attention behaviors in autistic children generalize before and after using the robot-mediated intervention systems.
    \item \textit{Limited interactions' contents:} the presented research study has reproduced a straightforward LTM-RI procedure in a limited number of sessions. However, repeatedly employing the same interaction content may eventually cause the ceiling effect (e.g., participants hit the target on the first prompt in most of the trials) and loss of interest (e.g., participants feeling tired of doing the same intervention again and again) after numerous sessions. Consequently, it could be beneficial to integrate new contents under the same interaction protocol, or even a radically different interaction protocol needs to be utilized to update and reinforce the training procedure.
    \item \textit{Single domain application:} although the authors have successfully validated LTM-RI for joint attention, no validation in other types of training has been presented. Therefore, the eventual value of LTM-RI will need to be also verified in different tasks and domains.
\end{itemize}
\section{Adaptive multi-robot therapies: introducing multi-agent communication and hybrid tasks for the treatment of ASD children}
\subsection{Introduction}
The fourth research we propose to analyze has been presented by Ali et al. and described in the companion paper \cite{ali2019adaptive}. Starting from the promising results achieved in recent years by applying social robotics to the medical treatment of ASD, the authors have proposed new fully autonomous intervention systems whose goal is to overcome the limitations of the previous ones. Specifically, the research work has made the following contribution to the state-of-the-art. Firstly, the focal point of the research is the formalization, design, and development of a single mathematical model for adaptive multi-robot-based therapy of autistic children for both LTM-based joint attention and imitation tasks. 
Secondly, the proposed system's effectiveness has been validated through the CARS scale. Thirdly, the research team has designed the first robot-mediated system to enhance multiple-person interaction capabilities for autistic people. The primary rationale for introducing inter-robot communication is that an autistic individual may also need to listen and look at others' communication in daily multi-communication settings. These multi-robots act as non-human partners to enhance the social communication abilities between multiple persons simultaneously. Moreover, the robots themselves act as therapists and stimulators for intervention without using any body-worn sensor during interventions. Based on the intervention results for the improvement of joint attention and imitation, it has proven to be the robot-mediated interventions (RMIs) as an evidence-based practice (EBP) in autism.
\subsection{Contribution to the state-of-the-art}
The main contribution to the state-of-the-art of the presented paper relies on the fact that the modules for joint attention and as well as imitation tasks are entirely adaptive. Several research studies have testified that the LTM-based prompt method is not restricted to imitation or joint attention tasks but can be used generally for any robot-mediated therapy. For instance, Zheng et al. \cite{zheng2015robot} have proposed a system to enhance the imitation capabilities of children. The child was asked to imitate the robot's gesture, and in case of failure, the robot points to improve the gesture. In another research study \cite{huskens2013promoting}, the robot therapy was based on asking open questions at the start, and if the child was unable to answer it correctly, then the robot adds a hint of the correct answer in it. Similarly, ARIA system \cite{esubalew2012step} applies the LTM-based protocol to improve joint attention of autistic children. However, none of these systems integrate a feedback loop that enables them to adapt to children's responses. 
In this respect, only one research has proposed a single robot-based adaptive model for improving joint attention only, but the system was unquestionably less sophisticated than the one presented in section \ref{txt:paper3sect}. The convenience of employing this model consists of the fact that it does not demand the constant engagement of the human therapist. In fact, it is difficult for each therapist to operate for extensive sessions with autistic children and robot-based therapies can also be administered at home. However, it is necessary to remember that non-human involvement has certain disadvantages (e.g., how to manage the situation if the child gets frustrated.)
\subsection{Trial's characteristic}
Two distinct and individual experiments form the intervention trial have been designed by the research team. Specifically, the intervention features a human-robot interaction without inter robot communication and a human-robot interaction with inter robot communication scenarios. These two procedures are presented as individual experiment architectures, and the focus of this paper is not to inter-compare the two of them since they have different goals.
\begin{itemize}
    \item \textit{Human-robot interaction without inter robot communication} The autistic child is subjected to an EEG before the intervention to measure brain activity and attentiveness. Consequently, the child is introduced to the first intervention, i.e., the LTM-based adaptive joint attention module. The child’s response is noted by NAO robots’ cameras. Subsequently to intervention completion, the child is newly subjected to an EEG room to measure brain state after therapy. The child is then introduced to the second intervention, i.e., imitation, which follows the same protocol for measuring brain activity before and after the therapy. In this latter therapy, as the child enters the intervention area, both the robots give stimuli by flashing their eyes (with the same color), after which the robots wait for the child to maintain eye contact with either robot for at least 5 seconds. Once eye contact is established, the robot is activated for imitation activity. In other words, this intervention basically employs the joint attention module so that imitation of the robot is activated by the eye contact of the autistic child.
    \item \textit{Human-robot interaction with inter robot communication} This experiment introduces the concept of inter robot communication along with human-robot interaction with the autistic child. This inter-robot communication is executed at the start of the experiment. As the intervention begins, both the robots are sitting and facing each other. One of them stands up and says \textit{"hello"} along with waving action to the partner robot. The partner robot responds by standing up, saying \textit{"hello"} coupled with a waving action. During this time period, the autistic child's response is registered as a listening task. Subsequently, a randomly selected robot turns towards the child and starts communicating similarly with him.  In return, the child's response is captured in terms of joint attention (i.e., child's eye contact), imitation (i.e., waving of hand) and speech. During this experimentation, it was also noted that children were paying attention to the robots and shifted gaze properly between robots during their inter-communication, as their typically developed peers.
\end{itemize}
\subsection{Prompts and Reward Provision System: The LTM Interaction Protocol}
\subsubsection{The Protocol's Structure}
\begin{figure}[!htbp]
    \centering
    {\includegraphics[width=1\linewidth]{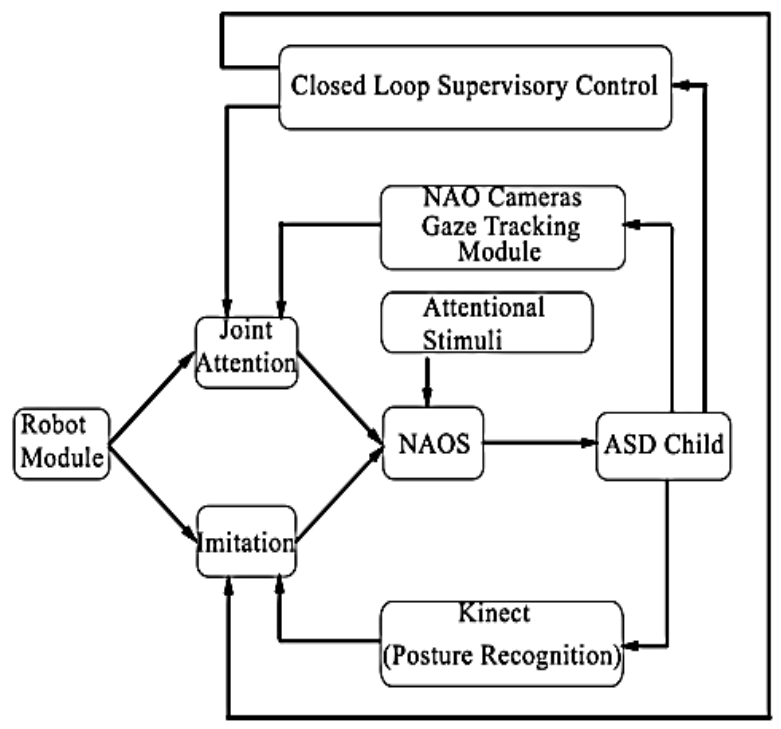}}
    \caption{Design protocol of joint attention and imitation module. The joint attention module is based on LTM protocol. Both of these modules use closed loop supervisory control.}
    \label{fig:im14}
\end{figure}

The interaction protocol for joint attention of MRIS employs least-to-most (LTM) cues, as shown in figure \ref{fig:im14}. In this respect, the logic underpinning the provision of different stimuli according to different prompt levels applies as described in the previous sections. \newline The protocol designed by the research team is based on three steps:
\begin{enumerate}[i)]
    \item \textit{Visual Cues:} the first protocol consists of visual cues. Specifically, two kinds of visual cues are available:
        \begin{itemize}
            \item "Rasta" (i.e., changing the eye color of the robot cyclically).
            \item "Blinking", considered the least intruding stimulus.
        \end{itemize}
        \item \textit{Speech cues:} the second protocol features speech cues combined with these visual cues. Speech cues added are \textit{"Hi"} and \textit{"Hello."} These speech cues are more effective hints for autistic children compared to visual cues \cite{libby2008comparison}.
    \item \textit{Motion cues:} the third protocol's level three comprises combined visual, speech, and motion cues. Specifically, the motion cues added are \textit{"move forward," "move backward," "stand-up," and "sit-down."} As previously pointed out, the cues are ordered according to the LTM approach \cite{zheng2017design}.
\end{enumerate}
%%%%%%%%%%%%%%%%%%%%%%%%%%%%%%%%%
%%%%%%%%%%%%%%%%%%%%%%%%%%%%%%%%%
\subsubsection{From the formal model to the algorithm}
\label{MRIS_LTM_formal}
Generally, proper intervention modeling requires various prompts, which are usually verbal and motion cues from the robot, along with some environmental factors. More precisely, the environmental factors are usually the medium towards which the robot points (e.g., the target monitors employed by Zheng et al. \cite{zheng2017design}). The present research study does not utilize any external factors.
The proposed system works as follows. The robot presents the prompt cues in LTM order, i.e., visual, speech, and motion cues, which differ in the level of complexity for eliciting the child’s expected response. Specifically, the model begins with the least intrusive cue for measuring the joint attention of the child. The libraries employed for representing different reinforcement stimuli are respectively \{V\}, \{S\}, and \{M\}, where \{V\} represents visual, \{S\} is for speech, and \{M\} represents motion. Combining these libraries makes it possible to obtain a set of different stimuli ranging from least to most in their effectiveness. For instance, using visual stimulus stand-alone has the least effect compared to using it in combination with speech and motion stimuli.
\subsubsection{Formalization of the model}
For the sake of completeness, we provide the reader the rigorous mathematical formalization of the MRIS-LTM model presented in the paper, along with its analysis. More precisely, the model is further explained by mathematically modeling its components: the joint attention and imitation modules, both formalized below. 
\hfill \break \indent
\paragraph{Joint attention module} \indent
With reference to the set of equations below, we provide a detailed explanation of the formal model for the joint attention module. This latter is indeed associated with the reinforcement stimulus employed by the robots to measure the joint attention of each subject. In fact, to execute this task, two modules are parallelly running under $O_{JA}$, a module to measure joint attention and a stimulus module. Equation $(4)$ represents parallel execution for true state through the \textit{AND} operation of the respective operands. While the first operand is related to joint attention recording for both robots, the second operand concerns reinforcement stimulus administered by the robots. Equation $(5)$ further explains the formal model. Specifically, $i$ denotes the robot number, $k$ denotes the number of eye contacts, and $j$ denotes the kind of reinforcement stimulus. $n$ and $m$ belong to real numbers. Moreover, equation $(6)$ denotes the different combination of stimuli, while equation $(7)$ represent the robot currently involved in the interaction.
\begin{equation*}
\begin{alignedat}{4}
    & S_1 = \textit{XOR}\{\textit{Initialization, Execution, Termination, Reward}\}\;(1)\\
    & S_2 = \textit{AND}\{Robot_1, Robot_2, GazeModule\}\hspace{2.3cm}(2)\\
    & S_3 = \textit{OR}\{V, V + S, V + S + M\}\hspace{3.75cm}(3)\\
    & O_{JA} = \textit{AND}\{\textit{Joint\_attention, Reinforcement\_Stimulus}\}\hspace{0.7cm}(4)\\
\end{alignedat}
\end{equation*}
\begin{equation*}
O_{JA} = \textit{AND}
\left\{
\begin{alignedat}{4}
 \sum_{k=1}^{n} \Big(\sum_{i=1}^{2}\Big(R_i, \int_{t=1}^{m} dt \Big), 
 \sum_{j=1}^{3} \Big(\sum_{i=1}^{2} (R_i,ST_j)\Big),\\
 \textit{where n,m $\in$ $\mathbb{R}$} \hspace{4.25cm} (5)\\
 \end{alignedat}
\right.
\end{equation*}
\begin{equation*}
    ST_{J} =
    \begin{cases}
      V, & \cond j=1 \\
      V + S, & \cond j=2 \hspace{4cm} (6)\\ 
      V + S + M, & \cond j=3
    \end{cases}
\end{equation*}
\begin{equation*}
    R_{i} =
    \begin{cases}
      \textit{Robot one}, & \cond i=1 \\
      \textit{Robot two}, & \cond i=2 \hspace{4.6cm} (7)
    \end{cases}
\end{equation*}
To further clarify equation $(5)$, we provide its illustration over multiple iterations:

\begin{equation*}
    \hspace{-0.5cm}
    \textit{$1^{st}$ iteration} =
    \begin{cases}
      (R_1,x_1) + (R_2,x_1), & \for j=1 \land i=1,2 \\
      (R_1,ST_1) \land (R_2,ST_1), & \for k=1 \land i=1,2 \\
    \end{cases}
\end{equation*}
\begin{equation*}
    \hspace{-0.5cm}
    \textit{$2^{nd}$ iteration} =
    \begin{cases}
      (R_1,x_2) + (R_2,x_2), & \for j=2 \land i=1,2 \\
      (R_1,ST_2) \land (R_2,ST_2), & \for k=2 \land i=1,2, \\
    \end{cases}
\end{equation*}
where $x_i$ denote the duration of the eye contact registered by the robot, while $ST_j$ indicates the kind of stimulus provided by the robots. The process described by the equations above continues until the completion of the therapy.

\paragraph{Imitation module} \indent
The MRIS imitation module's interaction protocol exploits the child's joint attention to activate the robot. Specifically, it has been allotted a specific time limit (5 seconds in the presented study) for which the child has to focus on the robot to activate it. The design choice of the threshold time ensures the activation of a single robot at a time and also makes the module adaptive. After proper eye contact is established with a specific robot, it starts its imitation tasks (i.e., it provides several prompts such as \textit{"Move Forward," "Move Backward," "Raise Hands," "Hands Down."} The depth-sensing camera (i.e., the Kinect) computes the success rate as the child imitates these gestures. It is also possible to involve a single robot in the imitation task experiment; However, introducing a second robot improves the child's multi-agent communication skills and imitation capabilities simultaneously, which becomes particularly useful in daily-life multi-agent scenarios. \newline As for the joint attention module, we provide a detailed explanation of the formal model for the imitation module.
\begin{equation*}
\begin{alignedat}{4}
    & S_1 = \textit{XOR}\{\textit{Initialization, Execution, Termination, Reward}\}\;(8)\\
    & S_2 = \textit{AND}\{Robot_1, Robot_2, GazeModule\}\hspace{2.3cm}(9)\\
    & S_3 = \textit{OR}\{ \textit{OR}\{\textit{Forward,Backward}\}, \\
    & \textit{OR}\{\textit{Raise hands, Hands down}\}\} \hspace{3.9cm}(10) \\ 
    & O_{JA\rightarrow IM} = \textit{AND}\{\textit{Joint\_attention, Imitation}\}\hspace{2.1cm}(11)\\
\end{alignedat}
\end{equation*}
The MRIS-LTM mathematical model for imitation based on joint attention is further illustrated by the mathematical equations represented in $(11)$ and $(12)$. Specifically, equation $(11)$ represents the imitation module based on the child's joint attention that triggers this module as the proper eye contact has been established. To execute this task, the joint attention module and the imitation module must run in parallel; hence it has been modeled through the \textit{AND} operation. The mathematical model expressed in equation $(11)$ is further described by equation $(12)$. In this latter, $O_{JA}\rightarrow IM$ symbolizes the output from joint attention integrated with the imitation module. Furthermore, $"i"$ denotes robot number, $"k"$ denotes the number of eye contacts and $"j"$ denotes the type of imitation, $"n"$ and $"m"$ belong to real numbers.
\begin{equation*}
%\hspace{-1cm}
O_{JA\rightarrow IM} = \textit{AND}
\left\{
\begin{alignedat}{4}
 \sum_{k=1}^{n} \Big(\sum_{i=1}^{2}\Big(R_i, \int_{t=1}^{m} dt \Big), 
 \sum_{i=1}^{2} \Big(\sum_{j=1}^{2} (R_i,IM_j)\Big),\\
 \textit{where n,m $\in$ $\mathbb{R}$} \hspace{4.0cm} (12)\\
 \end{alignedat}
\right.
\end{equation*}
Moreover, $IM_J$ represents the imitation sequence executed by robots and it is modeled as follows: 
\[ 
%\hspace{-0.5cm}
IM_J = \begin{cases}
    R_1 \rightarrow \begin{cases}
        \begin{alignedat}{2}
        \textit{forward} \\
        \textit{backward} 
        \end{alignedat}
    \end{cases}
    \hspace{0.35cm}\cond \textit{j=1}\\ \hspace{-0.35cm}
    \hspace{7.5cm} (13) \\
    R_2 \rightarrow \begin{cases}
        \begin{alignedat}{2}
        \textit{raise\_hands} \\
        \textit{hands\_down} 
        \end{alignedat}
    \end{cases}
    \cond \textit{j=2}\\
\end{cases}
\]

To further clarify equation $(12)$, we provide its illustration over multiple iterations:
\begin{equation*}
%\hspace{-0.35cm}
    \textit{$1^{st}$ iteration} =
    \begin{cases}
      (R_1,x_1) + (R_2,x_1), & \for i=1 \land j=1,2 \\
      (R_1,IM_1) \land (R_2,IM_2), & \for k=1 \land i=1,2 \\
    \end{cases}
\end{equation*}
\begin{equation*}
%\hspace{-0.35cm}
    \textit{$2^{nd}$ iteration} =
    \begin{cases}
      (R_1,x_2) + (R_2,x_2), & \for i=2 \land j=1,2 \\
      (R_1,IM_1) \land (R_2,IM_2), & \for k=2 \land i=1,2, \\
    \end{cases}
\end{equation*}
where $x_i$ denote the duration of the eye contact registered by the robot, while $IM_j$ indicates the imitation tasks performed by the robots. The process described by the equations above continues until the completion of the therapy's session.
\hfill \break \indent
\subsubsection{MRIS-LTM analysis: algorithm's breakdown}
\begin{figure}[!htbp]
    \centering
    {\includegraphics[width=1\linewidth]{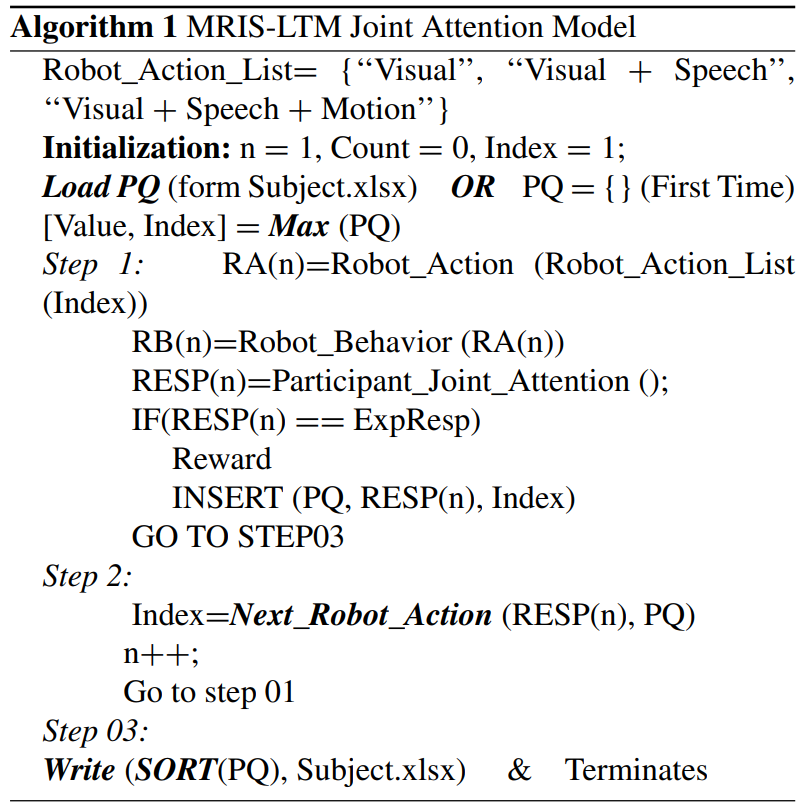}}
    \caption{Pseudocode of the algorithm derived from the formal model.}
    \label{fig:im15}
\end{figure}

With reference to figure \ref{fig:im15}, we analyze the proposed algorithm. The MRIS-LTM model is structured on the following steps: 
\begin{enumerate}[i)]
    \item \textit{Step 1:} firstly, the variable $Index$ is passed to $Robot\_Action\_List(.)$, which returns a robot action to be performed. The returned object is passed as input to the function $Robot\_Action(.)$, which provides the details of the action to be performed to the robot. At this point, the function $Robot\_Behavior(.)$ defines the behavior of the robot. Then $Participant\_Joint\_Attention(.)$ function begins. Specifically, it records the joint attention and gives an associated current response. If the current response matches with the expected response (i.e., $RESP(n)=ExpResp$), then the reward is given, the function \textit{INSERT(.)} records into a priority queue the obtained response and the index of the action, and the control is transferred to step 3. On the contrary, if the current response differs from the expected response, then step 2 activates, and step 1 is repeated until $Max\_Limit$ is reached (i.e., when $RESP(n) \neq ExpResp$ and \textit{$Max\_Limit=IN$}). Data is recorded, and code terminates.
    \item \textit{Step 2:} in the first place, it is checked if $Max\_Limit$ has been reached. If this is the case, the code terminates; otherwise, the current response is given to the $Next\_Robot\_Action(.)$ function, and step 1 is repeated until $PQ$ is filled or the child's response matches with the expected response.
    \item \textit{Step 3:} the code terminates after saving the sorted $PQ$ list into a file. $Max\_Limit$ represents the number of attempts available before the child's response matches the expected response.
\end{enumerate}
%%%%%%%%%%%%%%%%%%%%%%%%%%%%%%%%%
%%%%%%%%%%%%%%%%%%%%%%%%%%%%%%%%%
\subsection{Results}
\subsubsection{Results of the human-robot interaction without inter-robot communication}
\begin{figure}[!htbp]
    \centering
    {\includegraphics[width=1\linewidth]{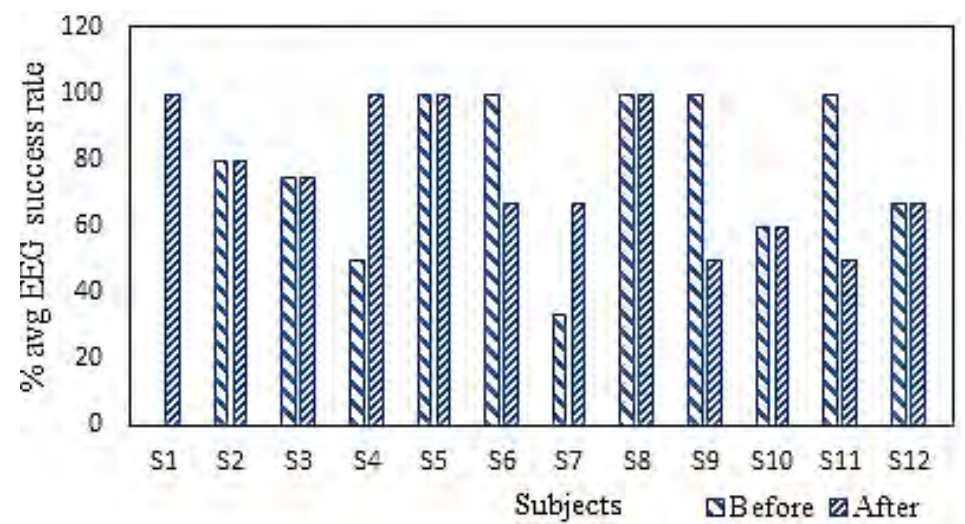}}
    \caption{Average EEG success rate of each individual before and after joint attention module.}
    \label{fig:im16}
\end{figure}

\begin{figure*}[!htbp]
    \centering
    {\includegraphics[width=1\linewidth]{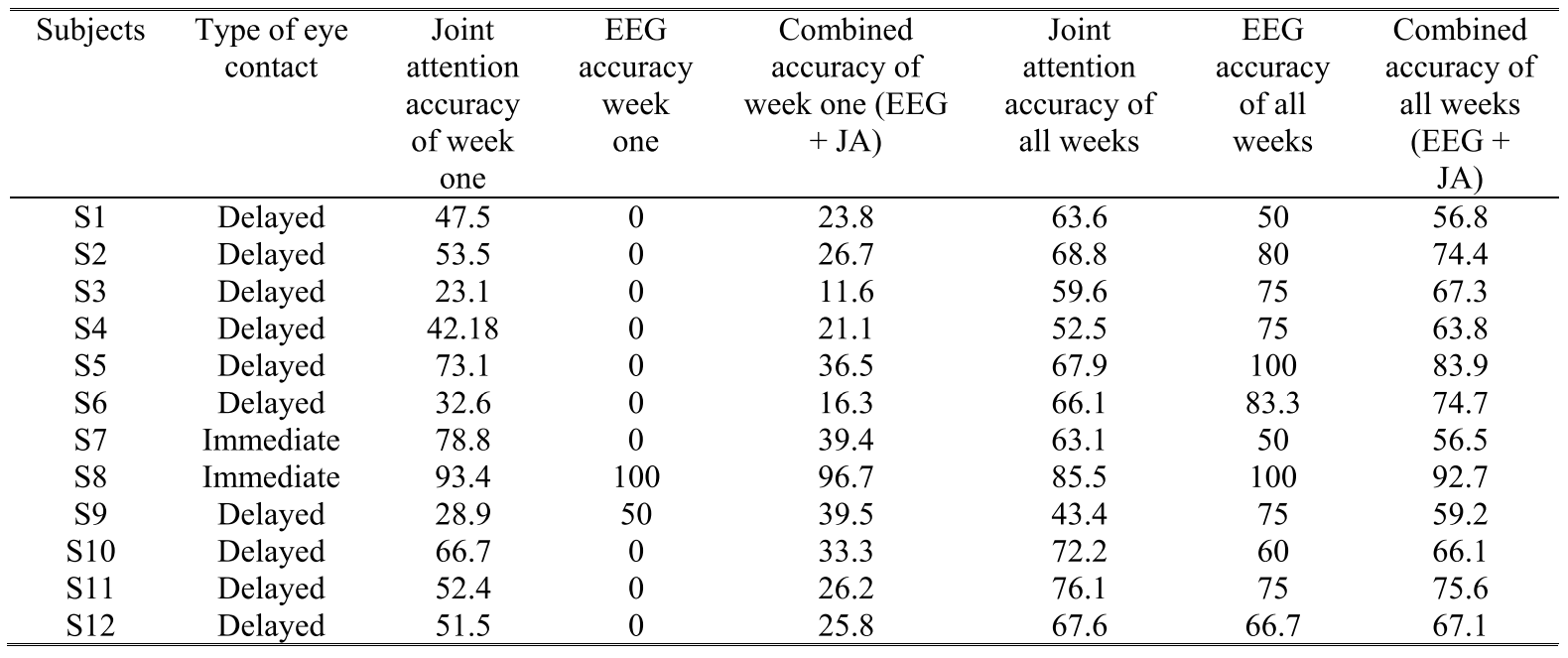}}
    \caption{Joint attention module's table.}
    \label{fig:table_att}
\end{figure*}

\begin{figure}[!htbp]
    \centering
    {\includegraphics[width=1\linewidth]{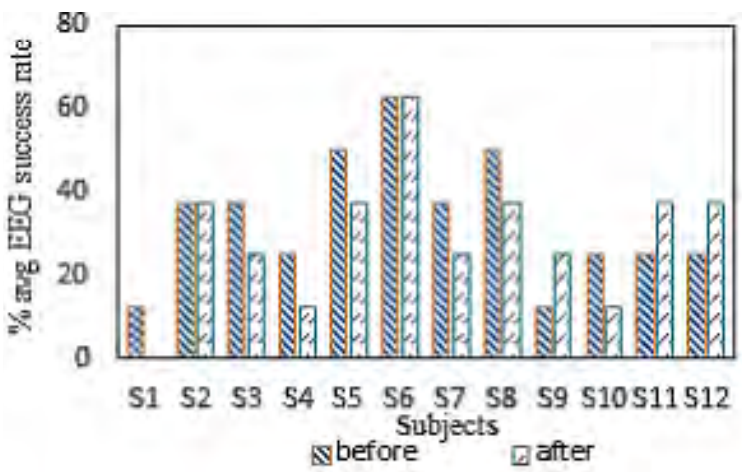}}
    \caption{Average EEG success rate of each individual before and after joint attention along with the imitation module.}
    \label{fig:im17}
\end{figure}

\begin{figure*}[!htbp]
    \centering
    {\includegraphics[width=1\linewidth]{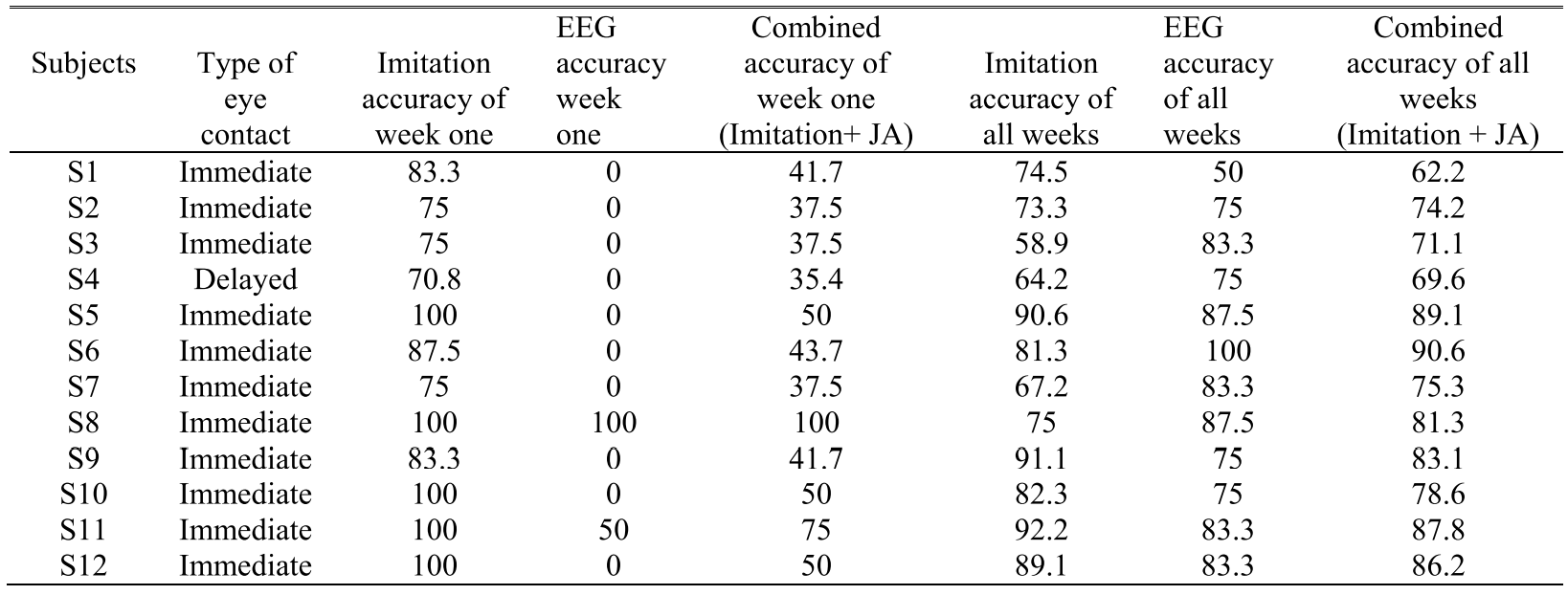}}
    \caption{Imitation module's table.}
    \label{fig:table_imit}
\end{figure*}

\begin{figure*}[ht]
    \centering
    {\includegraphics[width=1\linewidth]{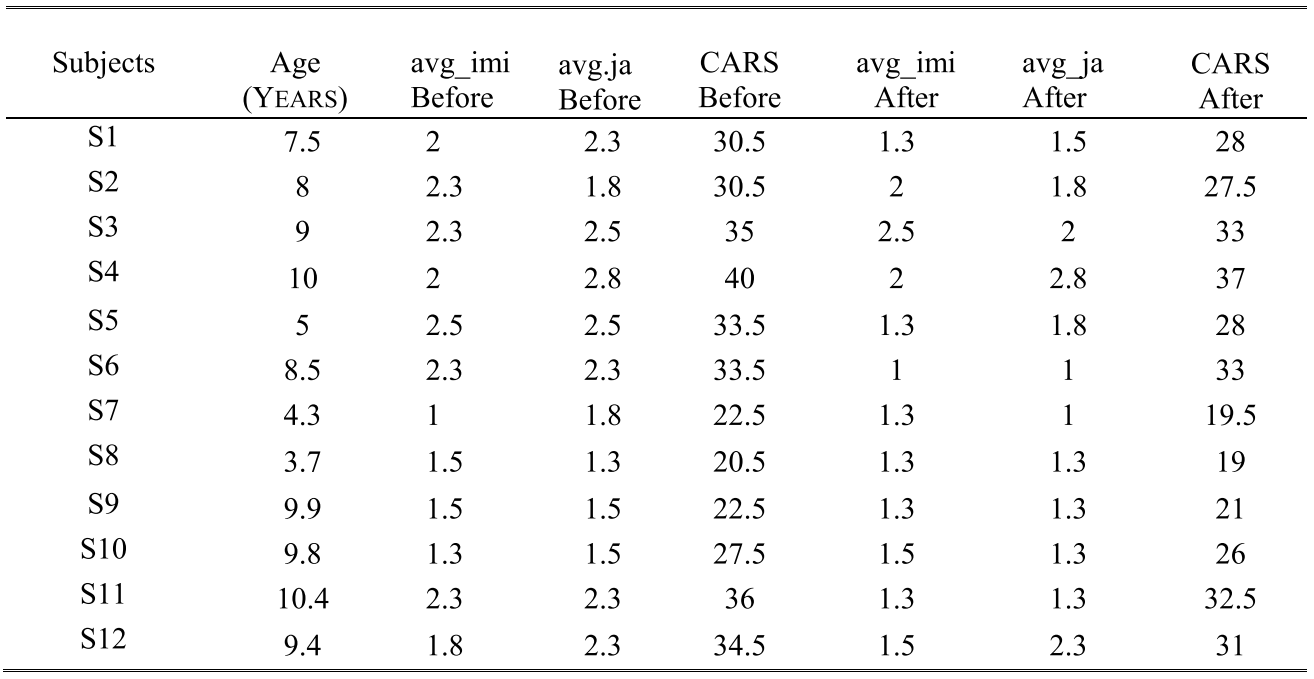}}
    \caption{Cars table for human-robot interaction with inter robot communication.}
    \label{fig:CARS_modules}
\end{figure*}

\begin{figure}[!htbp]
    \centering
    {\includegraphics[width=1\linewidth]{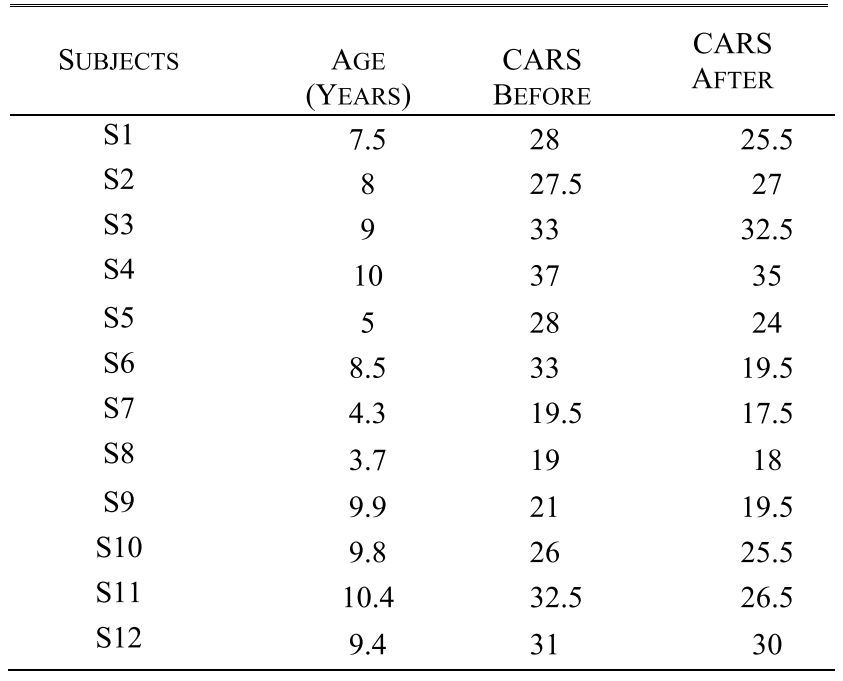}}
    \caption{ Cars table for human-robot interaction without inter robot communication.}
    \label{fig:CARS_global}
\end{figure}

\paragraph{Joint attention module's results}
The results of the joint attention module are based on two distinct criteria:  the eye contact of the autistic child that allows measuring the child's attentiveness towards a stimulus provided by a robot and the delay in shifting the gaze of the child from one robot to the other based on a given stimulus, which measures improvement in social and cognitive developments. Furthermore, the interest level of the participants in the joint attention module was measured before and after the intervention using EEG, as depicted in figure \ref{fig:im16}. Based on these criteria, figure \ref{fig:table_att} summarizes each child's behavior improvement in joint attention compared to the joint attention measured during the first week.
%%%%%%%%%%%%%%%%%%%%%%%%%%%%%%%%%
%%%%%%%%%%%%%%%%%%%%%%%%%%%%%%%%%
%%%%%%%%%%%%%%%%%%%%%%%%%%%%%%%%%
%%%%%%%%%%%%%%%%%%%%%%%%%%%%%%%%%
\paragraph{Imitation module's results}
The results of the imitation module are based on a single criterion, i.e., the imitation performed by the child when the robot gives a stimulus to measure the motor skills. Figure \ref{fig:im17} shows the results of the joint attention along with the imitation module. In fact, since the imitation module is activated through the joint attention module, their results are intrinsically related. Figure \ref{fig:table_imit} shows the overall improvement in each child's imitative behavior from week one. Furthermore, the results for both types of experimentation, i.e., joint attention and imitation modules, along with joint attention, have been verified using CARS score before and after the intervention, as presented in figures \ref{fig:CARS_modules} and \ref{fig:CARS_global}.\footnote{The Childhood Autism Rating Scale (CARS) is a widely-used behavior rating scale employed to help diagnose autism. Please note that the higher the CARS value registered, the more severe the patient's autistic traits are. } 
%%%%%%%%%%%%%%%%%%%%%%%%%%%%%%%%%
%%%%%%%%%%%%%%%%%%%%%%%%%%%%%%%%%
%%%%%%%%%%%%%%%%%%%%%%%%%%%%%%%%%
%%%%%%%%%%%%%%%%%%%%%%%%%%%%%%%%%
\subsubsection{Results of the human-robot interaction with inter-robot communication}
\begin{figure}[!htbp]
    \centering
    {\includegraphics[width=1\linewidth]{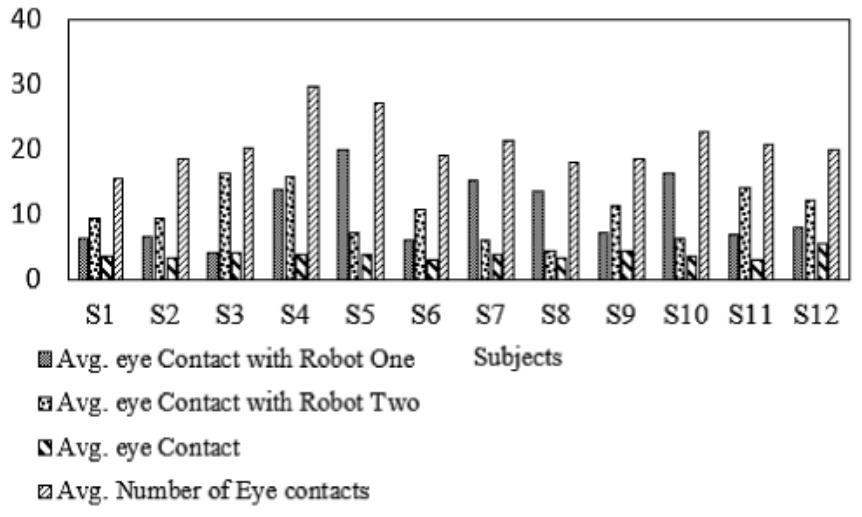}}
    \caption{Overall results for 12 ASD children for joint attention, imitation and response to inter-robot communication.}
    \label{fig:im18}
\end{figure}

Four different parameters have been considered to estimate the results of the human-robot interaction with inter-robot communication. Specifically, these parameters are the average eye contact with robots 1 and 2, the average eye contact in general, and the average number of eye contacts. Figure \ref{fig:im18} depicts the average values of these parameters for each subject. Furthermore, in this therapy, the attention paid to the robots' intercommunication and the waving and speech response of an autistic child towards both robots have been considered to estimate the therapy's effectiveness. 
%%%%%%%%%%%%%%%%%%%%%%%%%%%%%%%%%
%%%%%%%%%%%%%%%%%%%%%%%%%%%%%%%%%
%%%%%%%%%%%%%%%%%%%%%%%%%%%%%%%%%
%%%%%%%%%%%%%%%%%%%%%%%%%%%%%%%%%
%%%%%%%%%%%%%%%%%%%%%%%%%%%%%%%%%
%%%%%%%%%%%%%%%%%%%%%%%%%%%%%%%%%
\subsubsection{Observation on the results}
From the obtained results, it follows that this research study has three main contributions. Firstly, the research team has proposed, designed, and developed a single mathematical model for adaptive multi-robot-based therapy of autistic children for both LTM-based joint attention and imitation tasks. Secondly, the authors have validated the MRIS system's effectiveness by employing the CARS scale, which is the standard for these analyses. In doing so, it is possible to effectively understand if the therapy is beneficial and to what extent. Thirdly, this research study overcomes the limitations of earlier works by significantly improving the participants' multi-person interaction capabilities. \hfill \break \indent Concerning the proposed system's effectiveness, results prove that the eye contact duration of each participant has improved significantly over the experiments. Specifically, every participant showed some degree of improvement. Moreover, the delay in making eye contact with the robot after providing the stimulus has been reduced, i.e., the subjects became more responsive to the stimuli. For the imitation module, it has been observed that the participant activated both robots almost equally in recurring experiments. This latter result is crucial since it has proved the therapy to be successful for multi-interaction. However, it is necessary to mention that the percentage of success varied from child to child as each individual was responsive towards a different type of stimuli depending on his autism degree.
\subsection{Limitations}
This subsection's goal is to highlight the methodological limitations presented by this research study.
\begin{itemize}
    \item \textit{Small population sample and limited time frame of interaction:}
    One of the most impactful limits of this study is undoubtedly the small sample size considered. Although the achieved results show that employing the proposed system would be beneficial, validating it on a larger population sample is necessary. 
    \item \textit{Technological limitations:} In most cases, as previously pointed out, humanoid robots are not as capable of performing sophisticated actions yet. Specifically, though NAO is a state-of-the-art commercial humanoid robot, Kumazaki et al. \cite{kumazaki2018impact} have proved that it is not the optimal robot to employ for joint attention tasks. Furthermore, another significant technological limitation consists of the employment of EEG to measure children's brain activity and attentiveness. In fact, for EEG recording before and after interventions, some participants felt uncomfortable wearing EEG headsets. Therefore, their data has been excluded from the research. For this reason, it would be better to use another device instead of EEG as children are sometimes reluctant to wear body-worn devices.
    \item \textit{Generalization of the improved skill to human-human interaction:}
    Although several research studies have demonstrated the effectiveness of robotic stimuli and therapies for training autistic children in joint attention tasks, questions concerning the generalization of the acquired skill to human-human interaction, which is the ultimate goal of the therapy, remain open. Furthermore, the research community has not yet investigated whether the treatments' effects last in the long-term period. If these questions remain unanswered, there would not be a concrete application of social robotics in critical trials, nor every system, although particularly advanced, would not constitute a viable intervention tool.
\end{itemize}
\section{A comparative reading: introducing formal models to design profoundly structured and versatile interactions}
As previously stated, when two papers present a high degree of similarity in terms of their underlying logic, we provide the reader a comparative reading between their approaches. Since the main novelty offered by the last two papers analyzed concerns introducing two formal models for joint attention tasks, we propose a comparative reading of the algorithms derived from the formal models.
\subsection{Common Features}
\begin{itemize}
    \item \textit{The models' goal} Firstly, we note that both the papers have the same matrix, i.e., they propose a solution to solve the problem of unstructured and not standardized interactions. As previously pointed out in section \ref{txt:structuredInt_general}, structured learning environments enable therapists to help autistic people focusing on relevant stimuli and discern the most significant social signals among the many exchanged during social interactions. In this respect, both papers propose a specific prompt and reward provisions system that regulates the stimuli provided by the robot and establish a feedback mechanism that closes the loop, making the model adaptive to the participants' response.
    \item \textit{Rigorous but versatile models} It is worth mentioning that the need to design a formal mathematical model derives directly from the fact that it confers the interventions more effectiveness, reproducibility, and measurability. In fact, it is particularly challenging to measure the therapies' effectiveness without a standard reference or model. However, the rigor through which the models are formalized does not preclude their versatility. In fact, both the formal models proposed are natively applicable or adaptable to other tasks, with modifications in terms of contents and design choices. In this respect, these mathematical models play the role of operational approaches and are independent of the task to be executed. Furthermore, despite their definite structure, it is possible to adapt these models also to the execution of hybrid tasks, as demonstrated by Ali et al. \cite{ali2019adaptive}.
\end{itemize}
\subsection{Overcoming the Limitations of the First Formal Model}
The research work published by Ali et al. \cite{ali2019adaptive} can be considered the most sophisticated system proposed as of the time of this writing. In fact, it has overcome several limitations of the research study proposed by Zheng et al. \cite{zheng2017design}. The present subsection aims to clarify which limitations have been overcome and to what extent. For the sake of conciseness, we enlist them in the following items:
\begin{itemize}
    \item \textit{Introduction of multi-agent communication} The most significant limitation overcome by Ali et al. \cite{ali2019adaptive} concerns the introduction of multiple agents into the system. Specifically, during the second intervention, which involved inter-robot communication,  the child's behavior was documented when the robots were communicating with each other. This kind of protocol is typical in daily communication, hence the significance of introducing multi-agent communication into the systems. Furthermore, introducing a second robot also has a non-neglectable side effect because it enables the design of a complete learning environment. In other terms, previous research studies' interventions involved a single robot and were characterized by a single task to be executed (e.g., joint attention, imitation). However, a learning environment with this characteristic is very likely to be inadequate for extensive medical treatments (e.g., it may cause ceiling effects phenomena, loss of interest). In order to solve this issue, it is indeed necessary to design interventions that employ multiple agents (even more than two)  and propose the execution of hybrid tasks (e.g., joint attention and imitation in the same intervention). In doing so, it is possible to propose the autistic children a learning environment that is, on the one hand, structured and standardized and, on the other hand, characterized by a wide range of different stimuli.
    \item \textit{Employment of a proper eye-detection system for joint attention evaluation} Previous research studies (e.g., Zheng et al. \cite{zheng2017design}) have employed ALFace Detection system to determine whether the participants shifted their gaze correctly. Such a design choice constitutes unquestionably a significant drawback since an incorrect and not accurate evaluation metric may result in erroneous conclusions. Ali et al. \cite{ali2019adaptive} have overcome this technological limitation by employing ALGaze Analysis, which is a proper eye-detection system along with face detection. Furthermore, the system also features a depth-sensing camera (i.e., the Kinect) whose primary function is to evaluate the correctness of the participant's response during the imitation task. Nevertheless, the depth-sensing camera has also been integrated for head monitoring/face tracking to ensure the child's head's correct orientation during gaze shifting. In other terms, the Kinect's secondary purpose is to determine whether the gaze shifting results from a change of the participant's orientation of the torso or from a correct response to the provided prompt.
\end{itemize}
\subsection{Comparing the Respective Reward and Prompting Provision Systems}
The present section compares the previously analyzed two formal models and make some observations concerning their reward and prompting provision systems. %Subsequently, we propose the reader a novel algorithm, which has been derived directly from the present critical reading. Precisely, the suggested algorithm takes the best of both the presented algorithm and aims to overcome the limitations identified during the analysis. 
\begin{itemize}
    \item \textit{Prompts' complexity} With reference to the algorithm presented in section \ref{LTM_RI_formal}, it is possible to observe that the LTM-RI algorithm proposed by Zheng et al. \cite{zheng2017design} features a more elementary reward and prompts provision system based on the LTM protocol. This characteristic can be attributed to the fact that the task to be executed is not hybrid, i.e., it features only the joint attention module. Despite the basic prompts available, the algorithm reaches a higher degree of sophistication by integrating environmental factors into the prompts' generation process. %In doing so, the algorithm can generate a broader range of prompts.
    \newline At this point, we shift the focus of the analysis to the MRIS-LTM algorithm presented in section \ref{MRIS_LTM_formal}. With reference to its formalization, we recognize that the algorithm presents an advanced reward and prompts provision system that enables the generation of a wide variety of prompts. In fact, the availability of different combinations of visual, speech, and imitation stimuli supports the creation of an interactive and stimulating learning environment. However, it is worth mentioning that the research group has carried out a limiting design choice. Specifically, the team has decided not to integrate environmental factors in the formal model. These are generally identified as the objects the robot refers to by pointing actions, e.g., target monitors. %In this respect, a possible improvement to be made is to integrate environmental factors into the process of generating robots' behaviors. This betterment would be particularly suitable for the joint attention task since the act of shifting the child's attention toward a specific object is an effective strategy to empower children's joint attention's capabilities in social circumstances. Furthermore, the environmental factor can also be employed as a reward provision system, as proposed by Zheng et al. \cite{zheng2017design}. Specifically, if the child responded correctly to the prompt, the target monitor would display a brief video selected among the child's preferred cartoons.
    \item \textit{Prompts' level advancements} 
    \label{Prompts_level_advancements}
    A comparative reading of the previously analyzed models reveals an additional common characteristic between the two formal models. Specifically, it is possible to observe that the logic underpinning both the prompts provisioning systems in case the obtained response is different from the expected response (i.e., \textit{Resp $\neq$ ExpResp}) is the same. More precisely, we highlight that both the algorithms, derived from the corresponding formal models, envisage two possibilities if the child's response differs from the expected response:
    \begin{itemize}
        \item The prompts level remains the same, i.e., the following prompt that the system will propose will have the same level as the previous one.
        \item The prompts level increases, i.e., the following prompt that the system will propose will have a higher level than the previous one.
    \end{itemize}
    Both outcomes are perfectly consistent with the LTM paradigm's logic since the learner is subjected to more intrusive stimuli only in case of necessity. As previously pointed out in section \ref{txt:LTM_def}, this approach is based on the assumption that a combination of visual, speech, and motion prompts is more likely to stimulate the child's attention than a less intrusive prompt (e.g., eyes blinking). However, it is necessary to underline that heightening the prompts' level with no demonstrated need hinders, in certain respects, the learning process. %However, it is necessary to underline that heightening the prompts' level with no demonstrated need hinders, in certain respects, the learning process. In fact, since the child is less stimulated to respond correctly with the least intrusive prompt, the possibility of the prompt level remaining the same has been included. Specifically, this design choice has been carried out because the fact that the participant has not responded correctly to a prompt does not necessarily imply that it is required to heighten the prompt level to achieve the expected response. In fact, the attention capabilities of autistic children are, by definition, fluctuating; hence it is helpful to provide the participants a second opportunity to respond appropriately to the stimuli with the least intrusive prompt. 
    % \newline In this respect, it is possible to make another improvement to the algorithms presented. In light of the observations in this section, it would be beneficial to insert a counter variable for taking into account the number of prompts with the same prompt level the robot has provided and increase the prompt level only when a predetermined number of prompts with the same level have been provisioned. The logic underpinning this modification is the following. For instance, suppose to set the maximum number of attempts for each prompt level to 2. At this point, the robot provides a prompt with a specific prompt level; if the response obtained is not the expected one, the robot does not react immediately by increasing the prompt level but provides a prompt whose level is equal to the previous one, thus leaving the possibility to the participants to respond appropriately with the least intrusive prompt possible. In case the child cannot respond correctly on the second attempt, there are good chances that it will be necessary to increase the prompt level. In other terms, this modification aims to drive the robot to not increase the prompt level if not strictly needed, thus pushing the participants to stretch their joint attention/imitation capabilities. Obviously, the number of maximum attempts for each prompt level should be determined with the therapist, according to the child's autism severity, and, in any case, should not be excessively high since the interaction could become frustrating for the participant. 
    \section{A novel algorithm: introducing "Improved LTM-MRI algorithm"}
    The present section proposes a novel algorithm, which has been derived directly from the presented critical reading and aims to overcome the limitations identified during the analysis. The following items clarify the reasons that led to its development, summarize the rationale of applying the suggested improvements and provide the reader the intuition of its potential benefits.
    %The present section aims to present the proposed algorithm, clarify the reasons that led to its development, and provide the reader the intuition of its potential benefits. The following items summarize the rationale of applying the suggested improvements:
    \begin{itemize}
        \item \textit{Introducing interactive environmental factors} The first reason behind the proposal of the present algorithm is to overcome the limitation introduced by the design choice of not integrating environmental factors in the prompting system. In this respect, a possible improvement to be made is to integrate them into the process of generating robots' behaviors. This betterment would be particularly suitable for the joint attention task since the act of shifting the child's attention toward a specific object is an effective strategy to empower children's joint attention's capabilities in social circumstances. Furthermore, the environmental factor can also be employed as a reward provision system, as proposed by Zheng et al. \cite{zheng2017design}. Specifically, if the child responded correctly to the prompt, the target monitor would display a brief video selected among the child's preferred cartoons. Please note that the most effective intervention setup is built by introducing profoundly different environmental factors characterized by a solid interactive nature. In doing so, it is possible to design sophisticated prompting and reward systems according to the intervention's nature and the participant's needs. Collectively, the positive effect of employing environmental factors is two-fold. On the one hand, it provides the opportunity to generate varied but versatile forms of interactions, which could be tuned to the child's degree of cognitive impairment. On the other hand, it can also work as a practical and effective reward system to motivate children to maintain sustained motivation during interventions.
        \item \textit{Introducing a more sophisticated prompting and reward provision system} As previously observed in section \ref{Prompts_level_advancements}, both the reward prompting and reward systems are perfectly consistent with the LTM paradigm's logic. However, they tend to heighten the prompts' level with no concrete need, thus precluding the children from elaborating the least intrusive prompt and appropriately responding to it.
	    In this respect, it is worth mentioning that the attention capabilities of autistic children are, by definition, fluctuating. Once provided a prompt, the child requires a certain amount of time to process it, especially when inputs consist of not intrusive prompts, which are more difficult to elaborate correctly for autistic children.
        In other terms, children are less stimulated to respond correctly to the least intrusive prompt, but training children to respond to them maximizes the interventions' effectiveness in real-world scenarios. \newline For these reasons, it is helpful to provide the participants at least a second opportunity to respond appropriately to the stimuli with the least intrusive prompt. In this respect, the proposed algorithm provides for the possibility of the prompt level remaining the same after a failed attempt to respond. Specifically, this design choice has been carried out because the fact that the participant has not responded correctly to a prompt does not necessarily imply that it is required to heighten the prompt level to achieve the expected response. In fact, it could be the case that the child had not succeeded in elaborating the response in time, or his attention could be subjected to a temporary fluctuation. \newline
        To tackle this issue, the proposed algorithm includes the following improvement. In light of the observations in this section, it would be beneficial to insert a counter variable for taking into account the number of prompts with the same prompt level the robot has provided and increase the prompt level only when a predetermined number of prompts with the same level have been provisioned. The logic underpinning this modification is the following. For instance, suppose to set the maximum number of attempts for each prompt level to 2. At this point, the robot provides a prompt with a specific prompt level; if the response obtained is not the expected one, the robot does not react immediately by increasing the prompt level but provides a prompt whose level is equal to the previous one, thus leaving the possibility to the participants to respond appropriately with the least intrusive prompt possible. In case the child cannot respond correctly on the second attempt, it is likely that it will be necessary to increase the prompt level. In other terms, this modification aims to drive the robot to not increase the prompt level if not strictly needed, thus pushing the participants to stretch their joint attention and imitation capabilities. Obviously, the number of maximum attempts for each prompt level should be determined with the therapist, according to the child's autism severity, and, in any case, should not be excessively high since the interaction could become frustrating for the participant.
    \end{itemize}
    \subsection{Assessing the Algorithm's Effectiveness: \newline Trial Characteristics and Experimentation Scenarios}
    In order to assess the effectiveness of the proposed algorithm, we also provide the structure of a possible scenario where it could be tested. The proposed human-robot interaction protocol is inspired by the previously analyzed ones but is revised according to the novelties introduced by the algorithm. In particular, the experimental phase could be structured as follows. An autistic child is placed in the intervention area. The room must be furnished with a set of distributed interactive devices that differ in terms of the nature of provided stimuli and their intensity (e.g., target monitors producing audiovisual stimuli, devices that provide only visual or auditive stimuli). In doing so, it becomes possible to build a protocol whose stimuli escalate according to the participants' needs and autism's degree. During the interaction, the robot administers a LTM-based joint attention prompting protocol to the child, and a set of distributed cameras detect the child's response in terms of gaze direction. In this respect, it is necessary to employ a gaze detection module able to distinguish complete body rotations from proper gaze shiftings. Concerning the prompting system, a certain number of attempts must be set according to the participant's cognitive impairment. The algorithm allows the participant to respond correctly to an input with the same prompt level before increasing their intrusiveness. Depending on whether the interaction ends with a target hit, the robot provides an adequate reward and terminates the interaction, according to the protocol.
    \subsection{Deploying the Algorithm in Clinical Settings: \newline Expected Outcomes}
    The development and the consequent deployment of the proposed algorithm in clinical settings could bring several benefits. Precisely, the algorithm is expected to perform better than the previously presented ones for the following reasons:
    \begin{itemize}
        \item The algorithm exploits the more sophisticated prompting system to stimulate children to stretch their joint attention and imitation capabilities, thus providing functional training to tackle real-life social scenarios.
        \item The algorithm considers the participant's eventual attention fluctuations, which are, by definition, cardinal traits of autism spectrum disorders. In doing so, the suggested procedure does not increase the prompt level without actual need, leaving the opportunity for the participant to respond correctly to the least intrusive prompt possible.
        \item The possibility to generate multiple prompt inputs on different scales allows to build more articulated interactions with the following potential benefits:
        \begin{itemize}
            \item More diverse interactions encourage children to maintain sustained motivation in interacting with the robot. Moreover, it also contributes to avoiding loss of interest and ceiling effects, i.e., the scenarios in which children learn to reproduce the interaction instead of learning how to properly respond to the given inputs.
            \item Designing more sophisticated interactions also permits recreating real-life scenarios that train children to apply the acquired skills in realistic settings.
        \end{itemize}
    \end{itemize}

    \subsection{From Theory to Practice: Implementation Details} At the implementation level, the proposed modification could be realized as follows. There could be two counter variables that operate, respectively, locally and globally. The local counter starts from 0 and increases subsequently to any detected response. When the maximum number of attempts is reached (i.e., the obtained response is still different from the expected one), the prompt level is increased, and the local counter is set again to 0. On the other hand, the global counter increases only when the prompt level increases. The employment of this latter variable could be helpful to determine the number of attempts demanded by each specific participant to achieve a target hit (i.e., to provide a response that matches the expected one) and, most of all, to track its progress across multiple sessions.
\end{itemize}
%%%%%%%%%%%%%%%%%%%%%%%%%%%%%%%%%%%%%%%%%%%%%%%%%%%%%%%%%%%%%%%%%%%
\begin{PurelyF}
    \caption{\newline Improved Least-To-Most Multi-Robot-Interaction algorithm for the execution of hybrid tasks.}
    \begin{algorithmic}[1]
        \Statex{Robot\_Action\_List $=$ \{$RA_1,\; RA_2,\; RA_3$\}}
        \Statex{Env\_Factor\_List $=$ \{$EF_1,\; EF_2,\; EF_3$\}}
        \Statex{\text{$n_{max} = $ maximum\_prompt\_level}}
        \Statex{\text{$max\_attempts = $ maximum\_attempts\_for\_prompt\_level}}
        \Statex{\textbf{Initialization:}}
            \State{\text{Initial prompt level $=1$}}
            \State{\text{$n=1$}}
            \State{\text{Index$=1$}}
            \State{\text{Local\_counter$=0$}}
            \State{\text{Global\_counter$=0$}}
        \Statex{\textbf{\textit{Step 1: Initial prompt}}}
            \State{\text{RobAction(1) $=$ Robot\_Action(Robot\_Action\_List(1))}}
            \State{\text{Child\_Behavior(1) $=$ Behav\_det\_fun(Robot\_Action\_List(1), }}
            \Statex{\text{Env\_Factor\_List(1))}}
            \State{\text{Resp(1) = Behav\_eval\_fun(Child\_Behavior(1))}}
            \State{\text{$Local\_counter = Local\_counter + 1$ }}
            \IF{\text{$Resp = ExpResp$}}:
                \State{\text{Reward}}
                \State{\text{Go to Step 3}}
            \ENDIF
        \Statex{\textbf{Step 2: Iterative prompting loop}}
        \FOR{\text{Prompt level $n = 1 : n_{max}$}}
            \State{\text{[RobAction(n), EnvFactor(n)] = Prompt\_gen\_fun(Resp(n-1))}}
            \State{\text{Child\_Behavior(n) = Behav\_det\_fun(RobAction(n),}}
            \Statex{\text{\hspace{0.35 cm}EnvFactor(n))}}
            \State{\text{Resp(n) = Behav\_eval\_fun(Child\_Behavior(n))}}
            \State{\text{$Local\_counter = Local\_counter + 1$ }}
            \IF{\text{$Local\_counter == max\_attempts$}}:
                \State{\text{$Local\_counter = 0$ }}
                \State{\text{$Global\_counter = Global\_counter + 1 $ }}
                \State{$n = n+1$}
            \ENDIF
            \IF{\text{$Resp = ExpResp$}}:
                \State{\text{Reward}}
                \State{\text{Break}}
            \ENDIF
        \ENDFOR
        \Statex{\textbf{Step 3: Termination}}
        \State{\textbf{return} ($Global\_counter * 2 + Local\_counter$)}
        \State{Terminate\_Interaction}
    \end{algorithmic}
\end{PurelyF}
\noindent
%%%%%%%%%%%%%%%%%%%%%%%%%%%%%%%%%%%%%%%%%%%%%%%%%%%%%%%%%%%%%%%%%%%
\begin{center}
\begin{tabular}{ |p{2cm}|p{6cm}|  }
 \hline
 \multicolumn{2}{|c|}{Robot actions} \\
 \hline
 $RA_1$   & \textit{"Visual"}\\ \hline
 $RA_2$   & \textit{"Visual + Speech"}\\ \hline
 $RA_3$   & \textit{"Visual + Speech + Motion"}\\ \hline
\end{tabular}
\end{center}
\begin{center}
\begin{tabular}{ |p{2cm}|p{6cm}|  }
 \hline
 \multicolumn{2}{|c|}{Environmental factors} \\
 \hline
 $EF_1$   & \textit{"Target monitor displays a static picture"}\\ \hline
 $EF_2$   & \textit{"Target monitor displays an audio clip"}\\ \hline
 $EF_3$   & \textit{"Target monitor displays a video clip"}\\ \hline
\end{tabular}
\end{center}
With reference to the algorithm's formalization below, we specify the role of the employed functions in the following points:
\begin{itemize}
    \item Behav\_det\_fun is the function responsible for detecting the participant's behavior. Specifically, it takes as input a \textit{RA (RobAction)} and an \textit{EF(EnvFactor)} and, once detected the child's behavior, returns its representation. At the implementation level, it exploits a dedicated gaze tracking module, which extracts the participant's gaze direction (e.g., ALGazeAnalysis, as employed by Ali et al. \cite{ali2019adaptive})
    \item The behavior evaluation function \textit{(Behav\_eval\_fun)} takes as input the participant's behavior and determines whether the behavior matches the expected response \textit{ExpResp}. As previously observed, behaviors exist on a spectrum, and there may be some scenarios in which it is not possible to clearly define whether the obtained response matches the expected one. For the present case, we assume to be able to determine if \textit{Resp(n) $=$ ExpResp} or \textit{Resp(n) $\neq$ ExpResp}. At the implementation level, this function takes as input the child's behavior in response to a particular robot's action (e.g., the child's gaze and body directions) and evaluates whether the child has responded appropriately to the provided stimulus. For instance, if, once provided the prompt, the child rotates the whole body instead of shifting his gaze, this function has the role of evaluating if such a response could be considered a target hit.
    \item Prompt\_gen\_fun is the prompting generation function that, given a robot action \textit{RA} and an environmental factor \textit{EF}, generates a prompt and provides it to the participant. It is worth mentioning that a critical role in the prompt generation process is played by the prompt level that ensures that the learner is provided the least intrusive stimulus required by the child to perform \textit{ExpResp}. At the implementation level, \textit{Prompt\_gen\_fun} could be implemented as a sorted list of prompt levels following the LTM hierarchy. If the response detected is not the expected response, given \textit{RobAction$(n-1)=RA_i$} (i.e., robot action at time instance $n-1$) and \textit{$EnvFactor(n-1)=EF_j$} (i.e., environmental factor at time instance $n-1$), we extract the robot action for the next instant, \textit{RobAction(n)$=RA_l(l \geq i)$} , from \textit{\{RA\}} and the environmental factor for the next time instance, \textit{$EnvFactor(n)=EF_k (k \geq j)$}, from \textit{\{EF\}}. In doing so, the level of the next prompts depends on the value of the variable  \textit{Local\_timer}. %As previously pointed out, the algorithm also introduces two variables, denominated \textit{Local\_timer} and \textit{Global\_timer}. \newline
    \item \textit{Prompt generation procedure} The prompt generation procedure works as follows. An initial prompt is presented to the participant; if the obtained response matches the expected one, the reward is awarded, and the interaction terminates. On the contrary, if the collected response differs from the expected one, the variable \textit{Local\_counter} is increased, and another prompt \textit{with the same prompt level} is provisioned. If the participant does not respond to the prompt appropriately after a predetermined number of attempts, it is likely that the participant demands more intrusive stimuli to respond correctly. Hence, the prompt level is increased, and a new prompt \textit{with increased prompt level} (i.e., with greater informative content) is given. This process is repeated until the participant responds correctly or is no longer possible to provide further prompts (i.e., the robot has provided all the combinations of prompts of the maximum prompt level). When the interaction terminates, the algorithm returns the number of failed attempts the participant required before achieving a target hit.
\end{itemize} 
\section{Conclusions}
Joint attention is an early-developing social-communicative ability that plays a critical role in children's language and social development. Impaired joint attention development is a hallmark of children affected by autism spectrum disorders, making it crucial to target this skill in early intervention efforts. This paper reviews and analyzes several studies on the applications of social robotics to autism spectrum disorders, with a particular focus on joint attention tasks. In the first section, we defined autism spectrum disorders and their societal implications. The second section examined the necessity for technological aid and the potential of robot-assisted autism therapy. In the third section, we defined joint attention and its significance in children's neuro-developmental trajectories. The fourth and fifth sections provided detailed analyses of the research studies published by Warren et al. \cite{warren2015can} and Kumazaki et al. \cite{kumazaki2018impact}, respectively. We offered a comparative reading of previously examined works in the sixth section. The seventh and eighth sections presented an in-depth analysis of the studies by Zheng et al. \cite{zheng2017design} and Ali et al. \cite{ali2019adaptive}. The ninth section critically compared the formal models in sections VII and VIII. Lastly, the tenth section introduced a novel algorithm aimed at overcoming the limitations of the algorithms discussed in the previous sections. The following paragraphs aim to briefly draw conclusions concerning the application of social robotics to the medical treatment of autism spectrum disorders and discuss the necessary developments in this research field to deploy this technology in clinical settings effectively.
\subsection{Future Research Directions to Explore}
This paper aims to encourage the research community to develop a profoundly structured yet versatile interaction protocol to set a baseline standard. Establishing such a standard is crucial for several reasons. Children affected by autism spectrum disorders exhibit a wide variety of differences in terms of autism severity, symptom intensity, and special interests. This diversity hinders the measurability of the effectiveness of designed robot-mediated interventions. In other words, progress in developing social robots for autism therapy is complicated by the difficulty in measuring system effectiveness due to the absence of a baseline standard. Despite the structured nature of the standard, it should be broadly applicable and focus on crucial aspects such as the child's major impairments, intelligence quotient, and cultural background.
Another critical factor is facilitating collaborations between clinicians and roboticists. The most significant weakness of the reviewed studies is that the data tend to be qualitatively but not quantitatively rich. Experiments on robot-mediated interventions often involve descriptive case studies of a limited number of individuals. At present, no large-scale longitudinal studies provide quantitative measures of how autistic children respond to robot-assisted therapies. This situation is understandable, as developing and evaluating socially assistive robot systems for autism therapy requires joint work from researchers with diverse specializations, ranging from engineering to clinical research. Few research groups cover these disparate fields comprehensively, thus focusing on their particular strengths (e.g., robot design, interaction design, or evaluation). Without clinical psychiatrists and psychologists, most research teams lack long-term, continuous access to protected groups such as autistic children, making it challenging to measure the benefits of their design choices. Encouraging collaborations between robot designers and clinicians is therefore essential for promoting in-depth interaction studies.
Several significant research directions concerning engineering challenges have emerged from this review. One major challenge is developing advanced computer vision and actuation systems for robots engaging in imitative games. To induce imitation behavior in children, the robot must either be remotely commanded or programmed to emulate recognized physical behaviors autonomously. Thus, the robot must accurately detect the child's body motion, identify each component, estimate the child's body movement in three-dimensional space, and map these movements to its own, potentially limited, effectors to emulate them recognizably. Imitation is a complex research problem in robotics, even under simple, controlled conditions. Integrating detection, computation, and actuation into a robot suitable for autistic children represents a substantial engineering challenge. Another fundamental research direction involves developing robots capable of detecting and responding to users' actions and moods in real time. For example, a child sensitive to bright lights will not react well to therapies involving bright visual stimuli. Robots designed for autism therapies must recognize and adapt to such circumstances among and within children before being employed as autonomous entities in therapeutic interventions. Some research teams have proposed systems that recognize mental and emotional states from physiological \cite{feil2011automated} and behavioral \cite{liu2008online} data. However, further research is necessary before robotic systems can reliably adapt to patients' preferences and moods.
\subsection{Open Questions} Despite constant evolution in this research field, some fundamental open questions must be addressed to deploy social robots for autism therapy in clinical settings effectively. The first crucial question concerns the generalization of acquired abilities. The ultimate goal of robotic interventions is to enhance autistic children's core skills in deficit areas and enable them to apply these abilities in human-human interactions. Therefore, whether these skills generalize to human-human interactions rather than remaining confined to human-robot interactions must be investigated. The second significant question concerns the long-term effects of robot-mediated autism therapy. It should be explored whether robotic interventions ensure long-term benefits or if their effects are temporary. Comparing the long-term effects of robot-mediated therapies with those of well-established therapies provided by human specialists would be valuable. Another critical question is the necessity of regular interventions to reinforce acquired skills and prevent or repair regressions. Research should examine whether periodic interventions could also be robot-mediated to some extent or must be entirely conducted by human therapists. It would be helpful to design standardized protocols to determine the best setup for human and robot contributions and the frequency of sessions. However, as previously mentioned, a standardized protocol should serve as a baseline standard but be versatile enough to adapt to a subject's specific needs, considering symptom severity and the extent of the child's impairments.
\subsection{Concluding Remarks}
The research community has shown increasing interest in robot-assisted autism therapy over the past few years. This paper demonstrates that applying social robotics to the medical treatment of autism spectrum disorders is a particularly promising research area. Specifically, employing social robots to address joint attention impairments has proven to be a viable approach. However, further research is required to deploy robots effectively in clinical settings. Additionally, robot-mediated therapy appears to be suitable for addressing a wide range of core impairments associated with autism spectrum disorders. Several studies on robot-assisted autism therapy have confirmed its effectiveness in multiple areas, including communication (e.g., joint attention, imitation, engaging communication behaviors), emotion recognition, and developing appropriate sensitivity to physical contact.
Moreover, using remote robots for therapeutic interventions has been especially valuable during the COVID-19 pandemic or when direct interpersonal contact is impossible. However, the diverse and heterogeneous nature of autism spectrum disorders demands a high degree of adaptation and individualization in interventions. While robot-mediated therapies have numerous potential advantages, the current state-of-the-art robots for autistic children have not yet reached their full potential in clinical applications. Most studies use robots controlled by technical researchers, focusing on robot-child interactions rather than integrating robot-assisted interventions into care protocols and actual therapy or educational settings. This paper has addressed the potential benefits of employing robots in the therapy of autistic children as facilitators of social contact and as tools to enhance their capabilities in core deficit areas. It is crucial to foster a mutual understanding between healthcare professionals and robot developers in their joint mission to design effective robot interventions, enabling autistic children to overcome significant daily challenges.
\bibliographystyle{unsrt}
\bibliography{references}
\end{document}